\documentclass[12pt]{article}
\newcommand{\pmb}[1]{{\setbox0=\hbox{#1}%
		\kern-.025em\copy0\kern-\wd0
		\kern.05em\copy0\kern-\wd0
		\kern-.025em\raise.0433em\box0 }}

\usepackage{amssymb}
\usepackage{amsmath}
 
\usepackage{multirow}
\usepackage{latexsym}
\usepackage[pdftex]{graphicx}
\DeclareGraphicsExtensions{.pdf,.jpg}
\usepackage{epsf}
\usepackage{graphicx,epsfig}
\usepackage{epstopdf}
\usepackage{amssymb}
\usepackage{amsmath}
\usepackage{setspace}
\usepackage{color}
\usepackage{latexsym}
\usepackage{comment}
\usepackage[pdftex]{graphicx}
\usepackage[colorlinks,citecolor=blue]{hyperref}
\DeclareGraphicsExtensions{.pdf,.jpg}
\usepackage{multirow}
\usepackage{epsf}
\usepackage{array}
\usepackage{booktabs}
\usepackage{authblk}
\usepackage{float}
\usepackage{subfloat}
\usepackage{dblfloatfix}
\usepackage[none]{hyphenat}
\usepackage{microtype}
\DisableLigatures{encoding = *, family = *}
\usepackage{caption}
\usepackage{subcaption}

\def\thefootnote{${\dagger}$}\footnotetext{These authors contributed equally to this work.}

\makeatletter
\newcommand*{\rom}[1]{\expandafter\@slowromancap\romannumeral #1@}
\makeatother
\newcommand{\bey}{\begin{eqnarray}}
	\newcommand{\eey}{\end{eqnarray}}

\newcommand{\bec}{\begin{center}}
	\newcommand{\eec}{\end{center}}

\renewcommand{\theequation}{\arabic{equation}}
\begin{document}
	\pagestyle{myheadings}
	\setcounter{tocdepth}{2}
	\baselineskip22pt
	\belowdisplayskip11pt
	\belowdisplayshortskip11pt
	\renewcommand{\thefootnote}{\fnsymbol{hello}}
	\author[1]{Aisha Chandio ${\dagger}$}
	
\author[1] {Gong Gui${\dagger}$}
	\author[2]{Teerath Kumar${\dagger}$  }
	\author[3]{Irfan Ullah }
	\author[4]{ Ramin Ranjbarzadeh}
	
	\author[5]{Arunabha M Roy}
	\author[6]{Akhtar Hussain}
	\author[1]{Yao Shen
	\thanks{ Corresponding author.    Email addresses: yshen@cs.sjtu.edu.cn }}
	\affil[1]{\it Department of Computer Science and Engineering,
Shanghai Jiao Tong University, Shanghai, 200240, China.}
	\affil[2]{\it Department of Software Engineering, School of Computing, National University of computer and emerging sciences, Islamabad 44000, Pakistan}
	\affil[3]{\it School of Electrical Engineering and Computer Science (SEECS), National University of Sciences and Technology (NUST), Islamabad, 44000, Pakistan}
	\affil[4]{ \it School of Computing, Faculty of Engineering and Computing,
Dublin City University, Dublin ,Ireland.}
	\affil[5]{\it Aerospace Engineering Department, University of Michigan, Ann Arbor, MI 48109, USA}
		\affil[6]{\it Department of Information Technology, Quaid-e-Awam University of Engineering, Science, and Technology, Nawabshah,Pakistan}
\title{\large \bf Precise Single-stage Detector }
	\date{}
	\maketitle
	\bec
	\newpage
{\bf Abstract}
\eec
{\it Background and objectives:} Deep learning (DL) logarithms have shown an impressive performance in various tasks. 
 Among them, Single-stage object detectors (SSD)  mainly depends on classification network to extract features, multiple feature maps to predict, and classification confidence to guide the filtration of the overlapping prediction boxes. However, there are still two problems causing some inaccurate results: (1) In the process of feature extraction, with the layer-by-layer acquisition of semantic information, local information is gradually lost, resulting into less representative feature maps; (2) During the Non-Maximum Suppression (NMS) algorithm due to inconsistency in classification and regression tasks, the classification confidence and predicted detection position cannot accurately indicate the position of the prediction boxes.  {\it Methods:}
In order to  address these aforementioned issues, we propose a new architecture, a modified version of Single Shot Multibox Detector (SSD), named Precise Single Stage Detector (PSSD). Firstly, we improve the features by adding extra layers to SSD. 
Secondly, we construct a simple and effective feature enhancement module to expand the receptive field step by step for each layer and enhance its local and semantic information. Finally, we design a more efficient loss function to predict the IOU between the prediction boxes and ground truth boxes, and the threshold IOU guides classification training and attenuates the scores, which are used by the NMS algorithm.
{\it Main Results:} Benefiting from the above optimization, the proposed model PSSD achieves exciting performance in real-time. Specifically, with the hardware of Titan Xp and the input size of 320 pix, PSSD  achieves 33.8 mAP at 45 FPS speed on MS COCO benchmark and 81.28 mAP at 66 FPS speed on Pascal VOC 2007 outperforming state-of-the-art object detection models. Besides, the proposed model performs significantly well with larger input size. Under 512 pix, PSSD can obtain 37.2 mAP with 27 FPS on MS COCO and 82.82 mAP with 40 FPS on Pascal VOC 2007. The experiment results prove that the proposed model has a better trade-off between speed and accuracy.
	\\
	\\
\\
\\
	{\bf 1. Introduction :}
	\\
	\\
	In recent years, deep learning algorithms have  became a powerful tool which can automatically  capture
nonlinear and hierarchical features that  has shown great success on various applications, in particular, image domains   
such as  classification, segmentation, detection, captioning, and various others \cite{sharma2017review,sultana2020review,zou2019object,kumar2021class,kumar2021binary,jamil2022distinguishing}.
Furthermore, it has been extended for different classification tasks  including  
audio classification \cite{park2020search,kumar2020intra,park2020search,chandio2021audd, turab2022investigating}, text classification \cite{minaee2021deep,selva2021review,kowsari2019text,razno2019machine,aslam2021fake,ullah2021rweetminer}, various signals classification \cite{roycnn,roy2022efficient,roy2022multi,roy2022adaptive}, and multi-modality object classification  \cite{han2022multimodal,asvadi2018multimodal, aleem2022random},  event detection \cite{crocco2016audio,fu2010survey}, and various other applications \cite{khan2022introducing,roy2021finite,khan2022sql}. 
Among them, object detection \cite{jiang2022meanet,sindagi2019mvx,feng2020deep,roy2022fast,roy2022real,roy2021deep} has been a central interest to vast majority of  researchers. To this end, there are various algorithms such as  YOLO, fast-RCNN, faster-RCNN, and others that have been successfully used for object detection over the years. Deep learning algorithms for object detection has gained significant attention over the last few decades \cite{sharma2017review,sultana2020review,zou2019object}. Object detection, aiming at locating object instances from a large number of predefined categories in natural images, is one of the most basic and challenging problems in computer vision \cite{liu2021survey,ji2021cnn}.
With the rapid development of CNN, object detection has made remarkable progress and gradually turns into two main architectures: two-stage and single-stage. Two-stage algorithms such as Faster Recurrent Neural Network (FRCNN) \cite{ren2015faster,hiemann2021enhancement} where the first stage only distinguishes a large number of background regions and obtains rough object proposals without considering the specific class of the object. It has then followed by the second stage that classifies each proposal and optimizes the location according to the features extracted from CNN network \cite{pal2021deep,zaidi2022survey}. Due to the existence of refinement conducted by the second stage, two-stage algorithms cannot achieve performance in real-time \cite{zaidi2022survey, hiemann2021enhancement}. Therefore, single-stage algorithms have been the major priority for various object detection applications due to real-time detection \cite{liu2021survey,ji2021cnn}  and thus, it is the particular interest of the current work. The single-stage algorithms directly perform classification and location optimization based on the default boxes \cite{zaidi2022survey,hiemann2021enhancement}. For example, You Look Only Once (YOLO) \cite{redmon2016you} and SSD \cite{liu2016ssd} have achieved fast real-time detection speed but simultaneously sacrificed detection accuracy. In recent years, single-stage detectors are improving their accuracy, but still,  they cannot achieve a better trade-off between speed and accuracy \cite{hiemann2021enhancement}.

In this paper, on the premise of guaranteeing the real-time performance of the model,
we propose a novel architecture named Precise Single Stage Detector (PSSD)  based on the original SSD that   
 provides solutions to the two key questions: (1) How to enrich the information of features used by predictors without relying on a deep backbone of the model like ResNet-101~\cite{he2016deep}? (2) Whether it is reasonable to rely on classification confidence to determine the filtration of overlapping boxes in the process of the NMS algorithm? 
In the next subsections, we will address the aforementioned questions in order to establish the main challenges and corresponding improvements necessary that we have implemented in our proposed PPSD model. 
\\
\\
 {\bf 1.1  Feature richness:}
 \\
 \\
Considering the significant overhead caused by the image pyramid, SSD \cite{liu2016ssd} puts forward a feature pyramid to solve the problem of multi-scale detection. The deep features in the classification network contain more semantic information, which is suitable for identifying large objects, while the shallow low-level features are more suitable for small objects. However, the lack of semantic information in shallow features and the loss of local details in deep features deteriorate the precision of SSD. Feature Pyramid Networks (FPN) \cite{lin2017feature} adds deep semantic information to shallow features using a U-shaped structure to obtain a more effective feature pyramid, which improves the effect of small object detection. DetNet \cite{li2018detnet} combines dilated convolution to reduce local information loss by decreasing the down-sampling steps so that the positioning accuracy of large objects is improved. From the above-mentioned networks, it is noted that the features used in each scale predictor need not only suitable semantic information but also local texture information for more accurate positioning. The information richness of each level feature is significant to the detection effect. But the problem is how to construct a high-performance feature pyramid with as little overhead as possible.
\begin{figure}[h]
\centering
\includegraphics[width=1.0\columnwidth]{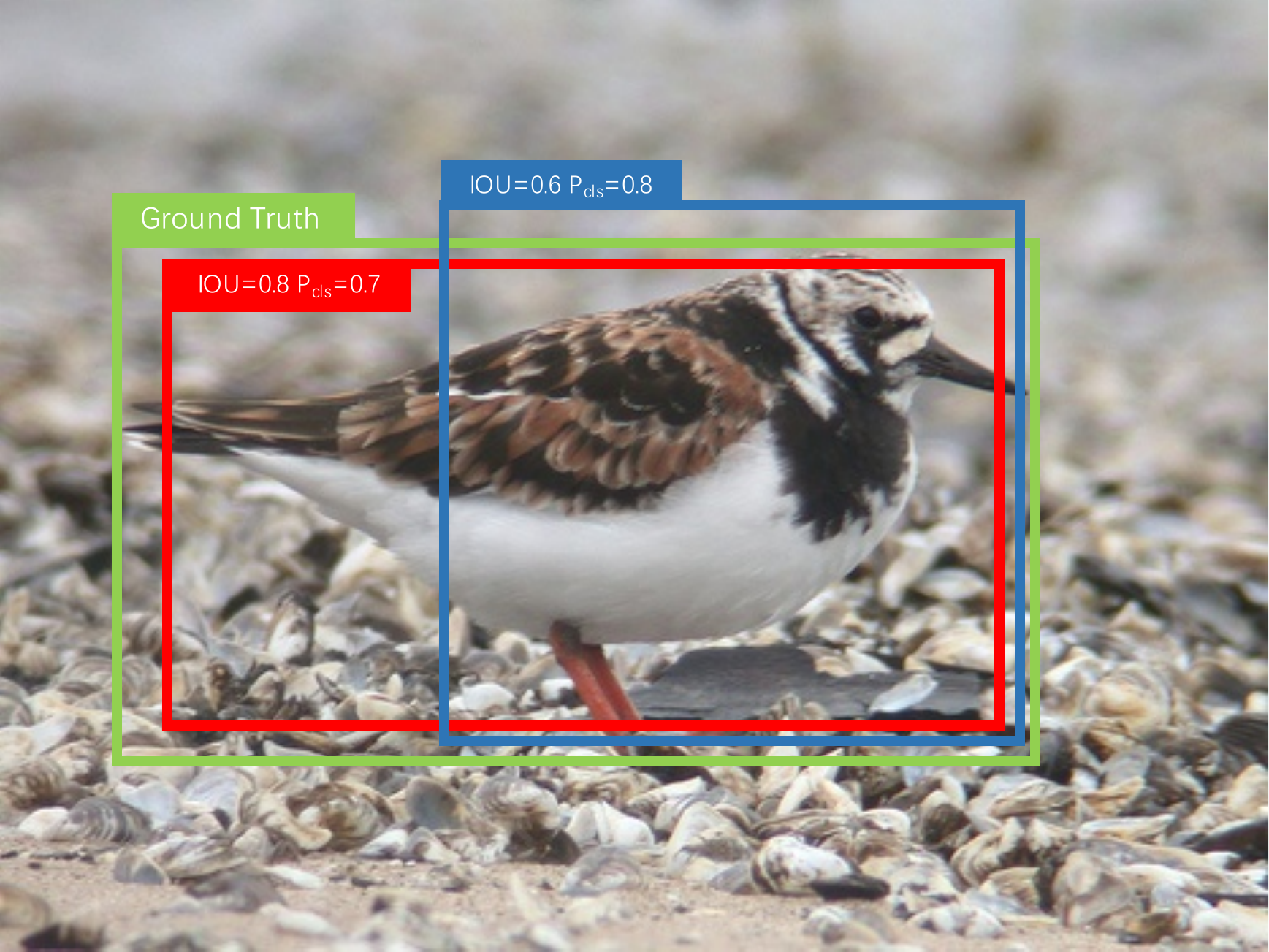} 
\caption{A example of test results without NMS, the blue box will be retained because of higher classification confidence.}
\label{fig1}
\end{figure}
\\
\\
{\bf 1.2 Filtration of overlapping boxes:}
\\
\\
Usually, to prevent the results from overlapping, we set the NMS as the final operation of object detection. In the NMS algorithm, the prediction boxes with the highest classification confidence are retained and other boxes are filtered when IOU between the two is greater than the threshold. As shown in Figure \ref{fig1}, this may lead to inaccurate results. IOUNet \cite{jiang2018acquisition} guides the NMS to alleviate this problem by predicting the IOU  between the regression boxes and their ground truth. Herein, the key question arises - how to make it more effective in single-stage algorithms?
\\
\\
{\bf 1.3 Contribution:}
\\
\\
To improve the detection effect, we mitigate the above problems based on the SSD model. Firstly, we improve features by introducing extra layers in the SSD to make the basic feature pyramid more suitable for feature fusion. Secondly, we send these features to the feature enhancement module (FEM). The FEM consists of two sub-modules, the receptive field expansion module (RFM) and two-way FPN. The RFM is used for expanding the receptive field of each feature step-by-step and two-way FPN is employed to supplement more local and semantic information. Finally, a new loss function is designed to predict the IOU between the regression boxes and their ground truth. Then we use the IOU label to guide the training of classification to improve the consistency of classification and regression tasks and use prediction IOU to guide inference to attenuate the scores of low-quality boxes.

We have validated  our method on MS COCO \cite{lin2014microsoft} and Pascal VOC \cite{everingham2010pascal} benchmarks. Under the hardware of Titan Xp with the input size of 320 pix, the proposed PSSD can obtain 33.8 mAP with 45 FPS on COCO and 81.28 mAP with 66 FPS on Pascal VOC 2007 outperforming state-of-the-art object detection models. Besides, our model also performs significantly well with a larger input size. Under 512 pix, PSSD can obtain 37.2 mAP with 27 FPS on MS COCO and 82.82 mAP with 40 FPS on Pascal VOC 2007. The experiment results prove that our model has a better trade-off between speed and accuracy.
The rest of this paper is organized as follows.: Section 2 covers  related literature relevant to this work; the proposed methodology has been described in section 3; the new loss function introduced in this study has been detailed in section 4; section 5 contains experimental results; 
 Section 6  deals with the relevant finding and discussion of the proposed model; the conclusions and prospects of the current work have been discussed in section 7.
\\
\\
{\bf 2. Related Work :}
\\
\\
{\bf 2.1 Object detection architectures:}
\\
\\
With the introduction of R-CNN \cite{girshick2015fast}, the deep learning method of object detection begin to develop which illustrated superior performance in object detection compared to traditional methods. To this end, various approaches such as DetNet \cite{li2018detnet}, Cascade R-CNN \cite{cai2018cascade} have been proposed. For each image, the R-CNN uses selective search \cite{uijlings2013selective} to create roughly 2000 region proposals. To provide more accurate candidate boxes of arbitrary sizes rapidly and limit the searching space in object recognition, the selective search method relies on simple bottom-up (BU) grouping and saliency cues \cite{deng2009imagenet, felzenszwalb2010object}. Each region proposal suggestion is twisted or cropped to a fixed resolution and the CNN network  \cite{krizhevsky2012imagenet} is used to extract the high-level, robust, and semantic 4096-dimensional feature as the final representation. Different region suggestions are assessed on a set of positive areas and negative regions using pre-trained category-specific linear SVMs for various classes. To obtain final bounding boxes for conserved item positions, the scored areas are modified using bounding box regression and filtered using greedy non-maximum suppression (NMS). As R-CNN obtained the fixed and same final representation for each region suggestion by cropping and warping, the item may exist in part in the cropped zone, and the warping procedure may result in undesirable geometric distortion. The accuracy of recognition will be harmed by these content losses or distortions, especially when object scales differ.

To mitigate this problem, SPP-NET \cite{he2015spatial} suggested a new CNN architecture called SPP-net based on the principle of spatial pyramid matching (SPM) \cite{lazebnik2006beyond}. SPM divides the image into a variety of divisions using several finer to coarser scales and accumulates these quantized local features into mid-level representations. Unlike R-CNN, SPP-net projects region suggestions of variable sizes to fixed-length feature vectors using feature maps from the fifth convolutional layer. The feature maps are reused because they include not only the intensity of local responses but also correlations with their spatial placements \cite{he2015spatial}. The SPP layer is the next layer following the final convolutional layer. Taking a three-level pyramid, if the number of feature maps in the convolutional layer is 256, the final feature vector for each suggested region acquired after the SPP layer has a size of $256 \times (12 + 22 + 42) = 5376$. Even though SPP-net outperforms R-CNN in terms of efficiency and accuracy, it still has several significant flaws. Feature extraction, network fine-tuning, SVM training, and bounding-box regressor fitting are all part of SPP-multistage net's workflow, which is nearly identical to R- CNN's. As a result, extra storage space is still necessary. Furthermore, the fine-tuning technique proposed in \cite{he2015spatial} cannot update the convolutional layers before the SPP layer. As a result, it's expected that the accuracy of very deep networks drops. Consequently, Girshick \cite{girshick2015fast} suggested a new CNN architecture called Fast R-CNN, which included a multitask loss on classification and bounding box regression. The entire image is processed with convolutional layers to yield feature maps, similar to SPP-Net. Then, from each suggested region with an ROI, a fixed-length feature vector is extracted. The RoI pooling layer is a subset of the SPP layer, which consists of only one pyramid level. Following that, each feature vector is fed into a series of FC layers before branching off into two sibling output layers. One output generates softmax probabilities for all the classes including the background while the second output layer uses four real-valued numbers to encode revised bounding-box coordinates. All parameters in these operations are optimized end-to-end using a multitask loss (except for the development of region suggestions). 
In order to build candidate boxes using biased sampling \cite{lenc2015r}, the latest object identification networks use various approaches to generate a set candidate pool of isolated region suggestions, such as selective search and Edgebox. The computation of region suggestion is also a hurdle in increasing efficiency. \cite{ren2015faster} proposed an extra region proposal network (RPN) that functions in a practically cost-free manner by sharing full-image convolutional features with the detection network to overcome this problem. RPN is accomplished using an FCN, which predicts both object bounds and scores simultaneously at every single position. Faster R-CNN \cite{ren2015faster} utilizes a network named Region Proposal Network(RPN) to generate region suggestions. Using anchors, this module learns to determine if an object is present at its associated position for each feature map. RPNs have a distinct and trainable module that predicts a set of bounding boxes considering reference boxes of various aspect ratios and sizes. As a result, this model defines a collection of anchors with varying aspect ratios and sizes on the relevant input images for each position on the feature extractor output. Although, on one hand, Faster R-CNN allows region proposal-based CNN architectures for object detection to be trained in an end-to-end manner, on the other hand, its training is highly time-intensive, and RPN creates objectlike regions together with the background rather than object instances, and it is not capable of coping with objects of extreme dimensions or forms. The R-FCN's last convolutional layer, Unlike Faster R-CNN, creates a total of $k^2$ position-sensitive score maps with a fixed grid of $k x k$ for each class. These are then appended to the position-sensitive RoI pooling layer’s output in order to get accumulative output. Finally, softmax responses across each class are generated by averaging $k^2$ position-sensitive scores in each RoI to form a $C + 1 - d$ vector. To achieve class-agnostic bounding boxes, another $4k^2-d$ convolutional layer is added. More powerful classification networks can be used with RFCN to achieve object detection in a fully convolutional architecture by sharing nearly all of the layers. 
In order to improve scale invariance, several object identification systems have used feature pyramids generated from the image \cite{felzenszwalb2010object, he2015spatial} at the cost of an increase in both training time and memory usage. Feature pyramid can extract rich semantics from all layers and be trained end to end with all scales. In a deep ConvNet, the in-network feature hierarchy provides feature maps with various spatial resolutions while also introducing huge semantic gaps due to varying depths.

To obtain a real-time detector, Single-stage architecture is investigated. SSD \cite{liu2016ssd} and YOLO series \cite{redmon2016you} are the most representative. However, their accuracies are lower than Two-stage algorithms. In recent years, Single-stage detectors have been improving, RetinaNet \cite{lin2017focal}, which uses more powerful classification networks, predictors, and dense anchors and proposes focal loss to alleviate the problem of several easy samples in the training process and enables Single-stage architecture accuracy to be aligned with Two-stage accuracy. In addition, in terms of lightweight Single-stage architecture, RFB-Net \cite{liu2018receptive}, RefineDet \cite{zhang2018single}, etc., improve detection accuracy by expanding the receptive field or learning more accurate default boxes. 
\\
\\
{\bf 2.2 Feature fusion:}
\\
\\
To solve the problem of multi-scale object detection, SSD \cite{liu2016ssd} proposes a feature pyramid. Then the feature fusion becomes popular. FPN \cite{lin2017feature} supplements the semantic information from top to bottom to improve the effect of small object detection. PA-Net \cite{liu2018path} builds another bottom-up feature pyramid based on FPN and adds horizontal connections to expand shallow and deep information for each layer. Kong et al \cite{kong2018deep}. propose a new efficient feature pyramid that integrates layer features in a highly nonlinear and efficient manner. RSSD \cite{jeong2017enhancement}, RFBNet \cite{liu2018receptive}, DSSD \cite{fu2017dssd}, etc. have also conduct feature fusion with different methods. In this paper, we propose a new architecture, a modified version of SSD, named PSSD by introducing extra layers. We use an efficient feature fusion structure to expand the receptive field of each layer and supplement its information, resulting in less redundancy and better results.
\\
\\
{\bf 2.3 IOU prediction:}
\\
\\
The drawback of using the classification confidence to filter the overlapping regression boxes is that a more accurate regression box is filtered out when its classification confidence is not the highest in the NMS algorithm. Based on the Two-stage architecture, IOUNet \cite{jiang2018acquisition} achieves exciting results by predicting the IOU confidence of the regression boxes and using them to guide the regression and NMS. We optimize the same issues in the single-stage architecture by designing a simple IOU-guided prediction structure.
\begin{figure*}[h]
\centering
\includegraphics[width=1.0\textwidth]{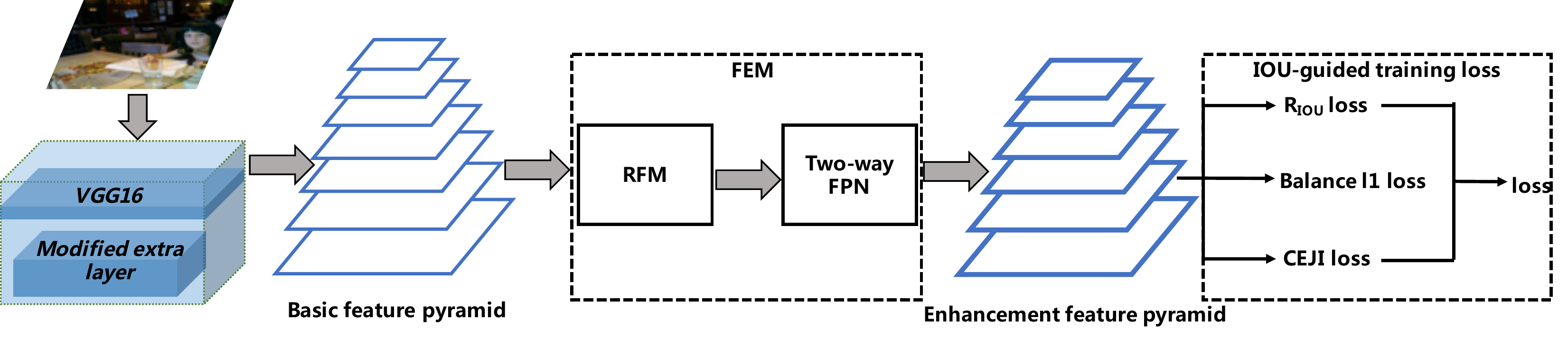} 
\caption{The overall network of the proposed PSSD with SSD backbone optimization, the introduction of a receptive field expansion module (RFM), implementation of two-way FPN and IOU-guided training loss to improve and optimize accuracy and detection speed.}
\label{fig2}
\end{figure*}
\\
\\
{\bf 3. Methodology} 
\\
\\
In the present study, our proposed PSSD model has been modified based on the original SSD model in order to improve and optimize the accuracy and detection speed. The overall proposed PSSD architecture has been shown in Figure \ref{fig2} and each modification has been elaborated in the following subsections.
\\
\\
\begin{figure*}[t]
\centering
\includegraphics[width=1.1\textwidth]{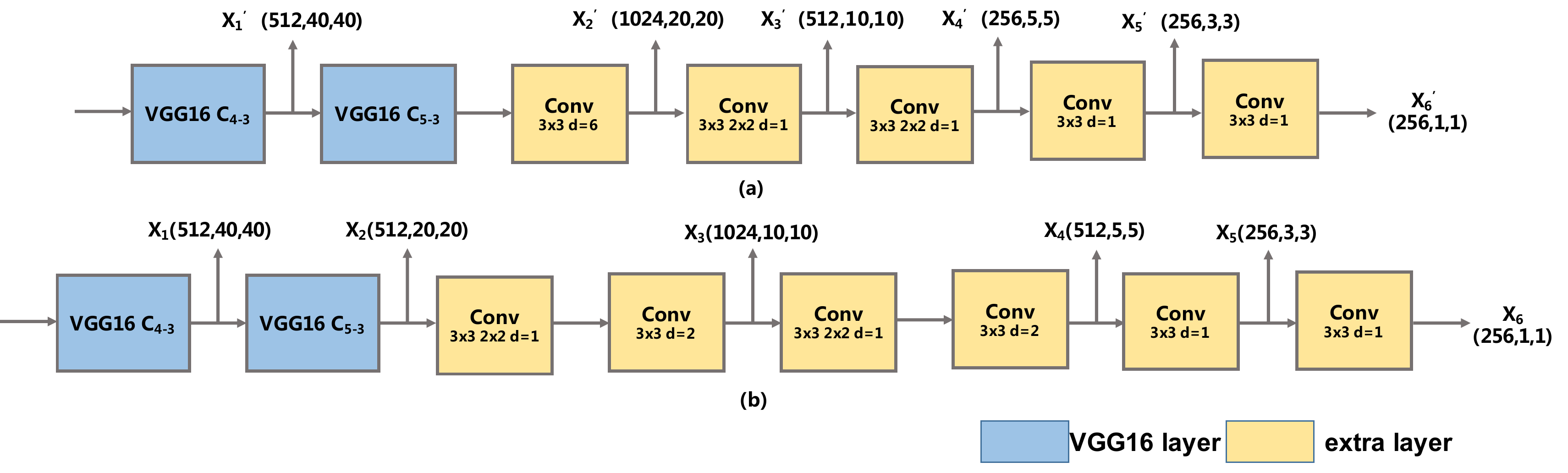} 
\caption{Enhancement of backbone layers (showing convolutional layers only) : (a) original extra layers of SSD; (b) our modified extra layers in PSSD for optimization of uniform basic feature maps distribution in the receptive field which is suitable for multi-scale feature fusion structures.}
\label{fig3}
\end{figure*}
\\
\\
{\bf 3.1 SSD backbone optimization}
\\
\\
The backbone of SSD consists of VGG16 \cite{simonyan2014very} and extra layers as shown in Figure \ref{fig3}. First of all, we reproduce the SSD algorithm and improve it with the following modifications: removing the L2 normalization of Conv$_{4-3}$ layer, adding BatchNorm \cite{ioffe2015batch} for each convolution layer, and replacing smooth $l_1$ with the balance $l_1$ \cite{pang2019libra} loss function. To improve the effect of small object detection, we have tried a four-layer FPN structure. However, the experimental results showed that the improvement is not as significant as expected. Thus, we have defined the feature maps which are obtained from the SSD backbone and used by predictors as the basic feature maps. The part of the SSD backbone is shown in Figure \ref{fig3} (a). The basic feature maps are $X_1^{'}$, $X_2^{'}$, $X_3^{'}$, $X_4^{'}$, $X_5^{'}$, $X_6^{'}$. To get $X_2^{'}$, $X_1^{'}$ performs three 3$\times$3 convolutions and one of 3$\times$3 dilated convolution with dilation of 6. The large dilation size leads to the loss of some of the object semantic and local information from $X_1$ to $X_2$. Meanwhile, the receptive field expansion of basic feature maps used by FPN is uneven. The expansion ratio of $X_1^{'}$ to $X_2^{'}$ is much larger than that of $X_2^{'}$ to $X_3^{'}$ and $X_3^{'}$ to $X_4^{'}$. Therefore, the effect of FPN is not apparent as expected. So, we redesign the extra layers under the following conditions.
\begin{itemize}
\item The receptive field of basic feature maps should expand evenly.
\item The parameters of extra layers without the pre-trained model should not increase drastically.
\item The coverage of the receptive field at each feature map should be appropriately expanded.
\end{itemize}
The new extra layers introduced in the proposed model have been shown in Figure \ref{fig3} (b). We use the 3$\times$3 convolution with dilation of 2 instead of two 3$\times$3 convolutions. In this way, the extension ratio of the receptive field from $X_2$ to $X_3$ and $X_3$ to $X_4$ are similar to $X_1$ to $X_2$ with a reduced number of network parameters. Considering the third aspect, we use the FEM module (see section 3.2 for details) to expand the receptive field of each feature map without affecting others. Overall, with this optimization, the receptive field distribution of basic feature maps is now more uniform and suitable for multi-scale feature fusion structures such as FPN.
\\
\\
\begin{figure}[t]
\centering
\includegraphics[width=1.1\columnwidth]{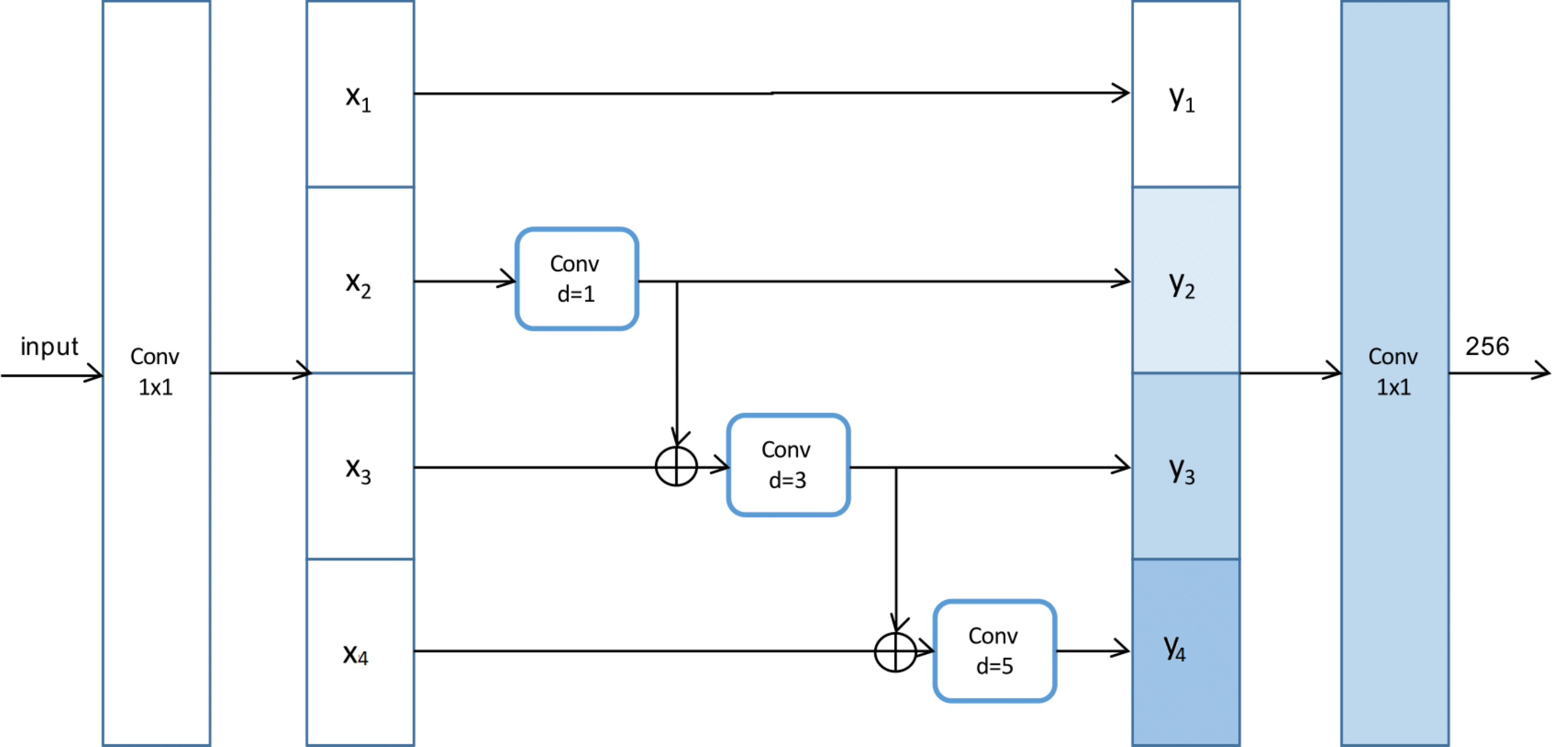} 
\caption{The proposed network structure of receptive field expansion module (RFM) for efficient feature enhancement in PSSD.}
\label{fig4}
\end{figure}
{\bf 3.2 Feature Enhancement Module}
\\
\\
{\bf 3.2.1 Receptive field expansion module:}
\\
\\
 The success of RFB-Net \cite{liu2018receptive} indicates that it is effective to use dilated convolution to expand the receptive field. Considering the effectiveness of the greater receptive field, we design a simple but efficient receptive field expansion module (RFM). As shown in Figure \ref{fig4},  to prevent large dilation size from influencing the calculation speed, we adopt the separation residual structure. First of all, we use 1$\times$1 convolution to perform feature conversion on the original feature map X, and then divide the X into four blocks: $X_1$, $X_2$, $X_3$, and $X_4$. Where $X_1$ maps directly to the final feature $Y_1$ to prevent information loss; $Y_2$ is gotten by operating a 3$\times$3, d=1 convolution on $X_3$, then add it with $X_3$ and send the sum to 3$\times$3, d=3 convolution get $_3$. Similarly, we conduct a 3$\times$3, d=5 convolution on the sum of $Y_3$ and $X_4$ to get $Y_4$. Finally, we use 1$\times$1 convolution to perform cross-channel information integration and dimension reduction on the extended feature Y. The separated residual design ensures that the receptive field of basic features is largely expanded in the case where the number of parameters does not increase significantly and the original feature details are kept as much as possible. This module is not expensive, except for the last layer, the input and output dimensions of each layer are the same \cite{ma2018shufflenet}.
\\
\\
{\bf 3.2.2 Two-way FPN:}
\\
\\
The design of the feature pyramid determines the feature richness and detection performance. FPN introduces high-level semantic information into a shallow layer to enrich features. However, the information on high-level features is also insufficient because of the loss of local details during feature extraction. Unlike DetNet \cite{li2018detnet}, RSSD \cite{jeong2017enhancement}, PANet \cite{liu2018path}, etc., which bring higher dimension or more convolution computation, we have implemented a simple two-way FPN to alleviate this problem. As shown in Figure \ref{fig5}, after RFM, we construct two information flows. Firstly, we introduce traditional FPN from the Y$_4$ feature to construct downward semantic information flows and get S$_1$, S$_2$, S$_3$, and S$_4$. Secondly, we construct upward local information flow from the shallower layer of VGG C$\_{3-3}$ to get L$\_1$, L$_2$, L$_3$, L$_4$, L$_5$, L$_6$. Finally, we combine the S features with the L features and use a 3$\times$3 convolution layer for feature conversion. To reduce overhead and retain the information of each feature, we use 256 dimensions to construct information flow, bilinear interpolation to do up-sampling, and Avg-pooling to do down-sampling. Finally, 512 dimension features are generated for training with stronger predictors.
\begin{figure}[t]
\centering
\includegraphics[width=1.1\textwidth]{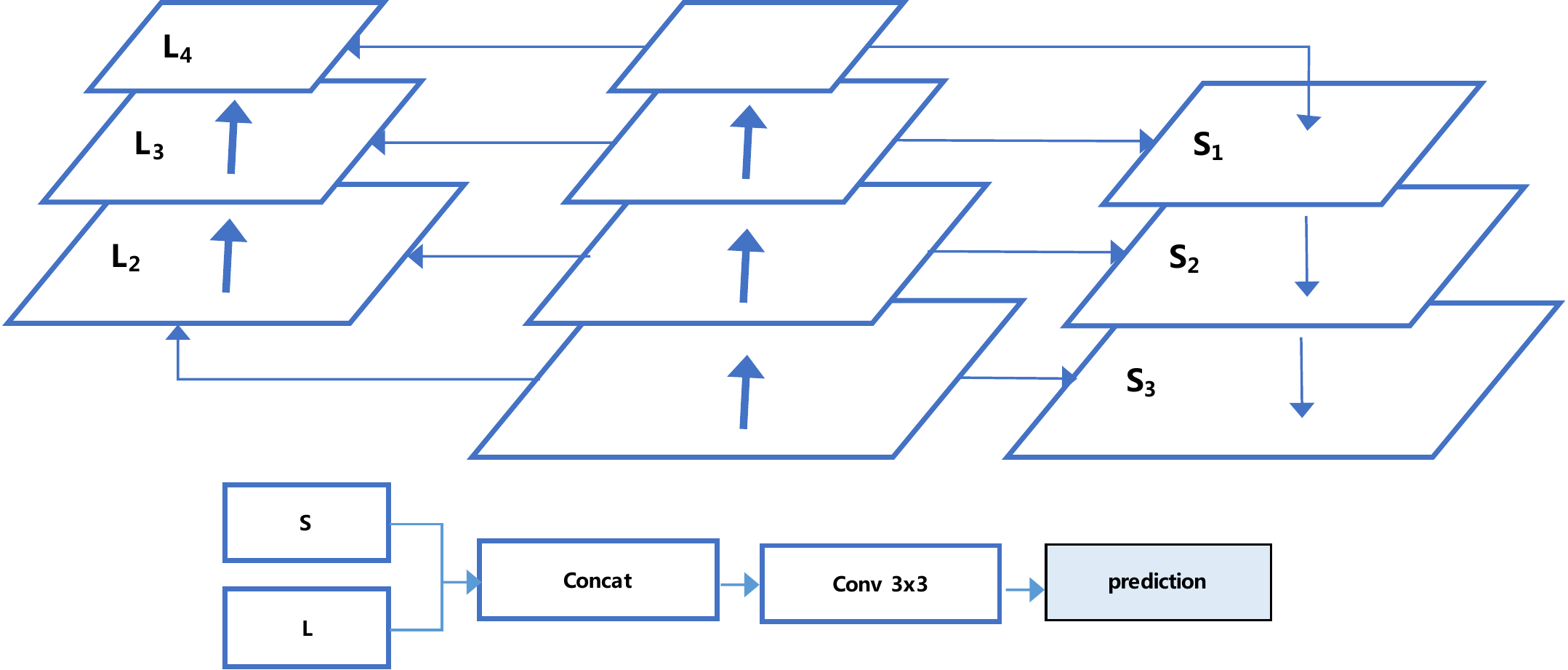} 
\caption{The network  structure of Two-way FPN in the proposed PSSD for enhancement of  feature richness and reduction of convolution computation for optimized detection performance.}
\label{fig5}
\end{figure}
\begin{figure}[hb]
\centering
\includegraphics[width=1.1\columnwidth]{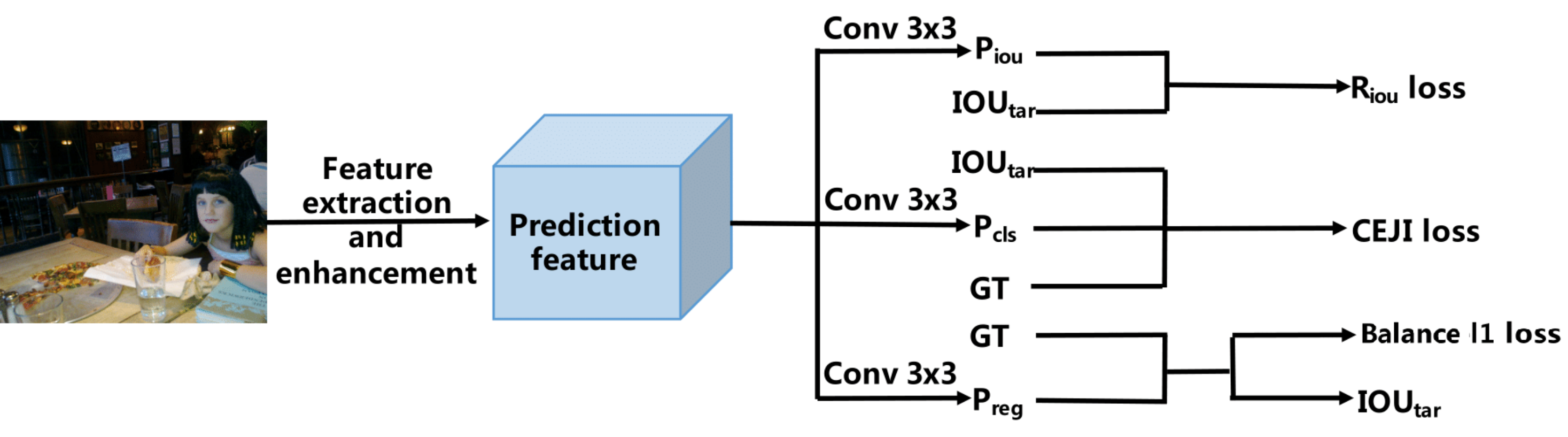} 
\caption{Proposed IOU-guided prediction structure in PSSD that consist of IOU regression loss ($ R_{IOU} loss$), cross-entropy of joint IOU ($CEJI loss$), and balance l$_1$ loss.}
\label{fig6}
\end{figure}
\\
\\
{\bf 4. Loss Settings based on IOU}
\\
\\
{\bf 4.1 IOU branch:}
\\
\\
The classification and regression tasks are relatively independent, classification confidence cannot accurately represent the predicted box positions. Hence, some more accurate prediction boxes are filtered by NMS. We use a less expensive category-independent way to do IOU-based attenuation for scores that are used in NMS. As shown in Figure \ref{fig6}, we add a branch to predict the IOU between the regression boxes and their ground truth. The results are gotten by directly applying 3$\times$3 convolution on the enhanced feature maps, and then using the sigmoid function to normalize it to 0 - 1. We calculate the IOU between the regression boxes and the ground truth. To ensure the convexity of loss and increase the gradient value, we design a new IOU regression loss ($ R_{IOU}loss$)  as follows:
$$ R_{IOU}loss=\left\{
\begin{aligned}
-\ln{\frac{P_{IOU}}{IOU_{tar}}} &      & {P_{IOU}      <      IOU_{tar}}\\
-\ln{\frac{IOU_{tar}}{P_{IOU}}} &      & {P_{IOU}      \geq      IOU_{tar}}\\
\end{aligned}
\right.
$$
where $P_{IOU}$ is predicted intersection over union and $IOU_{tar}$  is target intersection over union are precision of IOU and target IOU, respectively.
\\
\\
{\bf 4.2 Regression Loss:}
\\
\\
In order to make large objects regress better, we use balance l$_1$ ($ L_b(x)$) which proposed by Libra-RCNN \cite{pang2019libra}:
$$ L_b(x)=\left\{
\begin{aligned}
&\frac{a}{b}(b|x|+1){\ln(b|x|+1)}-\alpha|x|  & {|x|      <      1}\\
&\gamma|x|+C       & {otherwise}\\
\end{aligned}
\right.
$$
$$
\gamma=\alpha\ln(b+1)  
$$
where the default parameters are set as $\alpha$ = 0.5 and $\gamma$ = 1.5.
\\
\\
{\bf 4.3 Cross-entropy of joint IOU loss:}
\\
\\
For general Single-stage detection algorithms, such as RetinaNet \cite{lin2017focal}, the samples whose IOU is in the range of  0.4-0.5 are ignored, because it is difficult for the model to learn classification boundaries. However, the following problems still exist:
\begin{itemize}
\item There are some false positive samples (the IOU of regression results is less than 0.5),  and the model still forces them to be identified as objects, which results in low-quality prediction boxes.
\item There are some samples whose IOU are in 0.4 - 0.5 that can be learned, ignoring these samples is not conducive to improving the learning ability of the model.
\item The traditional cross-entropy loss does not take into account the generation of scores used in NMS during inference.
\end{itemize}
Thus, we adopt new positive sample rule and design a new cross-entropy of joint IOU ($CEJI(p,y,_{IOUtar}$) loss as follows:
$$ CEJI(p,y,_{IOUtar})=\left\{
\begin{aligned}
&-\ln(P_{cls}\cdot IOU_{tar})  & {IOU_{tar}      \geq      0.5}\\
&-\ln(P_{cls})       & {IOU_{tar}      <      0.4}\\
\end{aligned}
\right.
$$
Where {$IOU_{tar}$} is a gradient value, $P_{cls}$ is the predicted class probability. we set default boxes whose IOU with their ground truth of more than 0.4 as positive samples. If the default boxes can be regressed more accurately ({$IOU_{tar}\geq0.5$}), we calculate the classification loss, otherwise ignore them. Therefore, the model does not force the classification of fake positive samples and can excavate more potential positive samples. Meanwhile, the final optimization goal is changed to maximize {$P_{IOU}\cdot P_{cls}$}, which is consistent with the inference ({$scores=P_{IOU}\cdot P_{cls}$}). Additionally, considering that balance l$_1$ is not optimized from the overall of the regression boxes, this loss function can add a monitoring signal about the IOU to regressors to make the prediction results more accurate.
\\
\\
{\bf 5. Experiments}\label{sec:experim}
\\
\\
We conduct experiments on the MS COCO \cite{lin2014microsoft} and Pascal VOC \cite{everingham2010pascal} datasets. We use Cudnn10, two RTX TITANs to train the models, and a Titan Xp to test the results. The pre-trained parameters of VGG16 and VGG16$_{bn}$ are public. Our experiments include the following sections:
\begin{itemize}
\item Introducing the details of the experiments;
\item Demonstrating comparisons of PSSD with other advanced state-of-the-art methods;
\item Extensive experiment of PSSD model and demonstrating the validity of each sub-module.
\end{itemize}
{\bf 5.1 Performance evaluation on Microsoft COCO:}
\\
\\
At first, in order to evaluate the performance of PSSD, we perform experiments on the MS COCO dataset that contains 80 categories, 117266 training images, 5k verification images, and 20k test images. This experiment is conducted on the MS COCO \emph{2017\,train} set. We set the batch size to 32, the size ratio of default boxes to 0.06, 0.15, 0.33, 0.51, 0.69, 0.87, 1.05, the weight decay to 0.0005 and the momentum of SGD to 0.9. During training,  the initial learning rate is set to 0.002, which is reduced by 10 times subsequently at both the 90th and 120th epochs. The total number of training epochs is prescribed as 150. Other settings preserve the original SSD strategy. We conduct the test on MS COCO \emph{2017\,test-dev} set whose labels are not public and use the coco server to evaluate results.
\begin{table*}[h]
 \caption{
Performance comparison including detection accuracy, detection speed (in FPS) for various input sizes between state-of-the-art models and proposed PSSD considering different backbones evaluated  on MS COCO \emph{2017\,test-dev} set. \newline}
\resizebox{1.1\textwidth}{!}
{
\begin{tabular}{c|c|c|c|ccc|ccc}
\hline  
\multirow{2}{*}{Method} & \multirow{2}{*}{Backbone} & \multirow{2}{*}{Input size} & \multirow{2}{*}{FPS} & \multicolumn{3}{|r|}{Avg. Precision, IOU:} & \multicolumn{3}{|r}{Avg. Precision, Area:}\\
&&&& 0.5:0.95 & 0.5 & 0.75 & \quad S & \quad M & L \\
\hline  
Faster R-CNN \cite{ren2015faster} & VGG16 & $\sim$1000$\times$600 & 7.0 & 24.2 & 45.3 & 23.5 & \quad 7.7 & \quad 26.4 & 37.1 \\
Faster R-CNN w FPN \cite{lin2017feature} & ResNet-101 & $\sim$1000$\times$600 & 6.0 & 36.2 & 59.1 & 39.0 & \quad 18.2 & \quad 39.0 & 48.2 \\
R-FCN \cite{dai2016r} & ResNet-101 & $\sim$1000$\times$600 & 9 & 29.9 & 51.9 & - & \quad 10.8 & \quad 32.8 & 45.0 \\
CoupleNet \cite{zhu2017couplenet} & ResNet-101 & $\sim$1000$\times$600 & 8.2 & 34.4 & 54.8 & 37.2 & \quad 13.4 & \quad 38.1 & 50.8 \\
Deformable R-FCN \cite{dai2017deformable} & Inc-Res-v2 & $\sim$1000$\times$600 & - & 37.5 & 58.0 & 40.8 & \quad 19.4 & \quad 40.1 & 52.5 \\
\hline
\hline
SSD300 \cite{liu2016ssd} & VGG16 & 300$\times$300 & 43 & 25.1 & 43.1 & 25.8 & \quad 6.6 & \quad 25.9 & 41.4 \\
DSSD321 \cite{fu2017dssd}  & ResNet-101 & 321$\times$321 & 9.5 & 28.0 & 46.1 & 29.2 & \quad 7.4 & \quad 28.1 & 47.6 \\
YOLO$_{v3}$320 \cite{redmon2018yolov3} & DarkNet53 & 320$\times$320 & 45 & 28.2 & - & - & \quad 51.5 & \quad - & - \\
RefineDet320 \cite{zhang2018single}  & VGG16 & 320$\times$320 & 38.7 & 29.4 & 49.2 & 31.3 & \quad 10.0 & \quad 32.0 & 44.4 \\
RFB-Net300 \cite{liu2018receptive}  & VGG16 & 300$\times$300 & 66.7 & 30.3 & 49.3 & 31.8 & \quad 11.8 & \quad 31.9 & 45.9 \\
\textbf{PSSD320}  & VGG16 & 320$\times$320 & \textbf{45} & \textbf{33.8} & \textbf{52.2} & \textbf{35.8} & \quad \textbf{14.8} & \quad \textbf{38.5} & \textbf{50.3} \\
\hline
\hline
SSD512 \cite{liu2016ssd} & VGG16 & 512$\times$512 & 22 & 28.8 & 48.5 & 30.3 & \quad 10.9 & \quad 31.8 & 43.5 \\
DSSD513 \cite{fu2017dssd} & ResNet-101 & 513$\times$513 & 53.3 & 35.2 & \quad 13.0 & \quad 35.4 & 51.1 \\
YOLO$_{v3}$608 \cite{redmon2018yolov3} & DarkNet53 & 608$\times$608 & 19.8 & 33.0 & 57.9 & 34.4 & \quad 18.3 & \quad 35.4 & 41.9 \\
RefineDet512 \cite{zhang2018single} & VGG16 & 512$\times$512 & 22.3 & 33.0 & 54.5 & 35.5 & \quad 16.3 & \quad 36.3 & 44.3 \\
RFB-Net512 \cite{liu2018receptive}  & VGG16 & 512$\times$512 & 33.3 & 33.8 & 54.2 & 35.9 & \quad 16.2 & \quad 37.1 & 47.4 \\
\textbf{PSSD512}  & VGG16 & 512$\times$512 & \textbf{27} & \textbf{37.2} & \textbf{55.9} & \textbf{40.3} & \quad \textbf{18.7} & \quad \textbf{41.6} & \textbf{51.4} \\
\hline
\end{tabular}
}
\hspace*{\fill} \\
\label{table1}
\end{table*}
\begin{figure}
    \centering
    \subcaptionbox{}{
    \begin{minipage}[b]{1.0\linewidth}
    \includegraphics[width=1\linewidth]{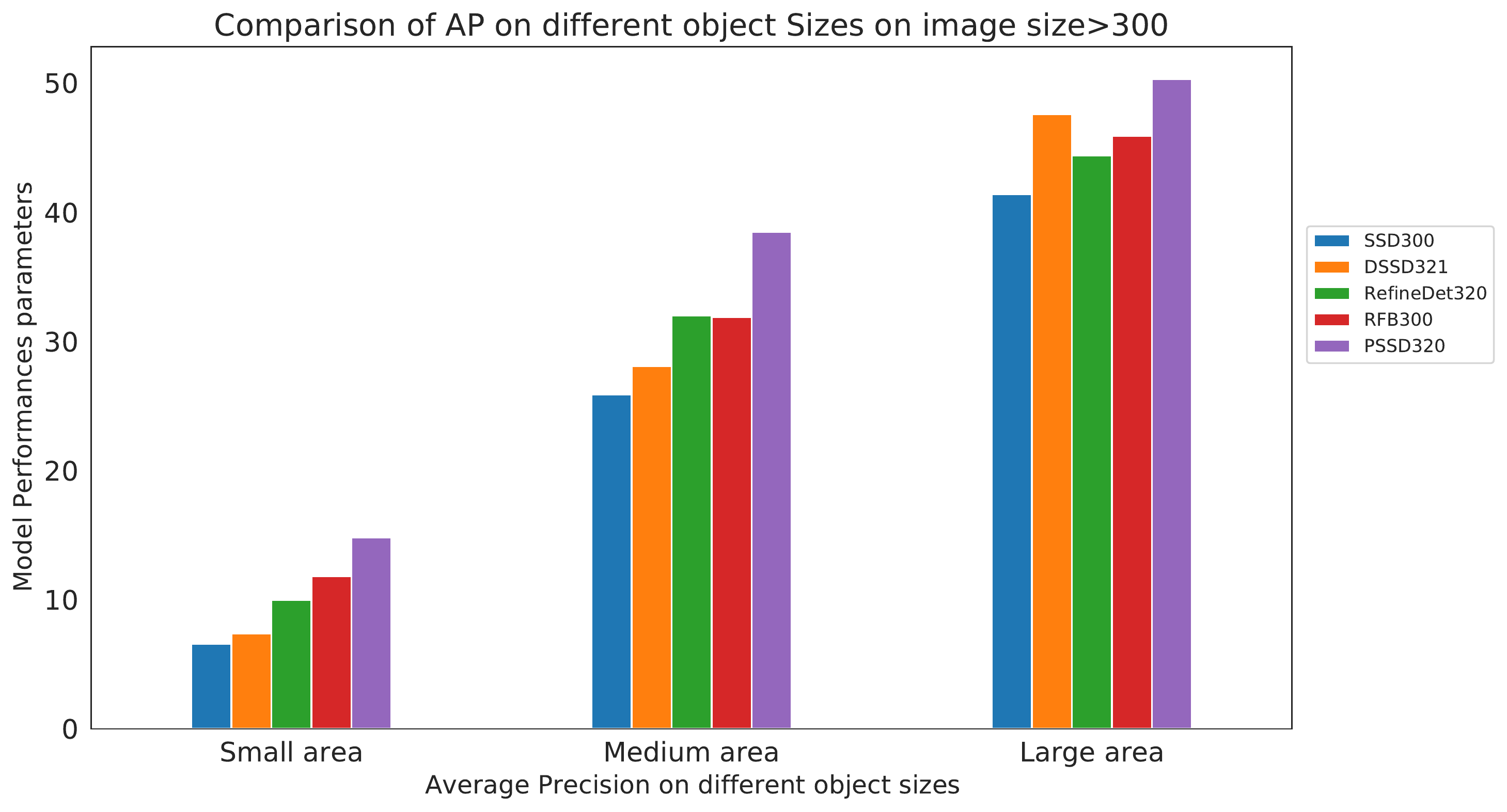}
    \end{minipage}
    }
    \subcaptionbox{}{
    \begin{minipage}[b]{1.0\linewidth}
    \includegraphics[width=1\linewidth]{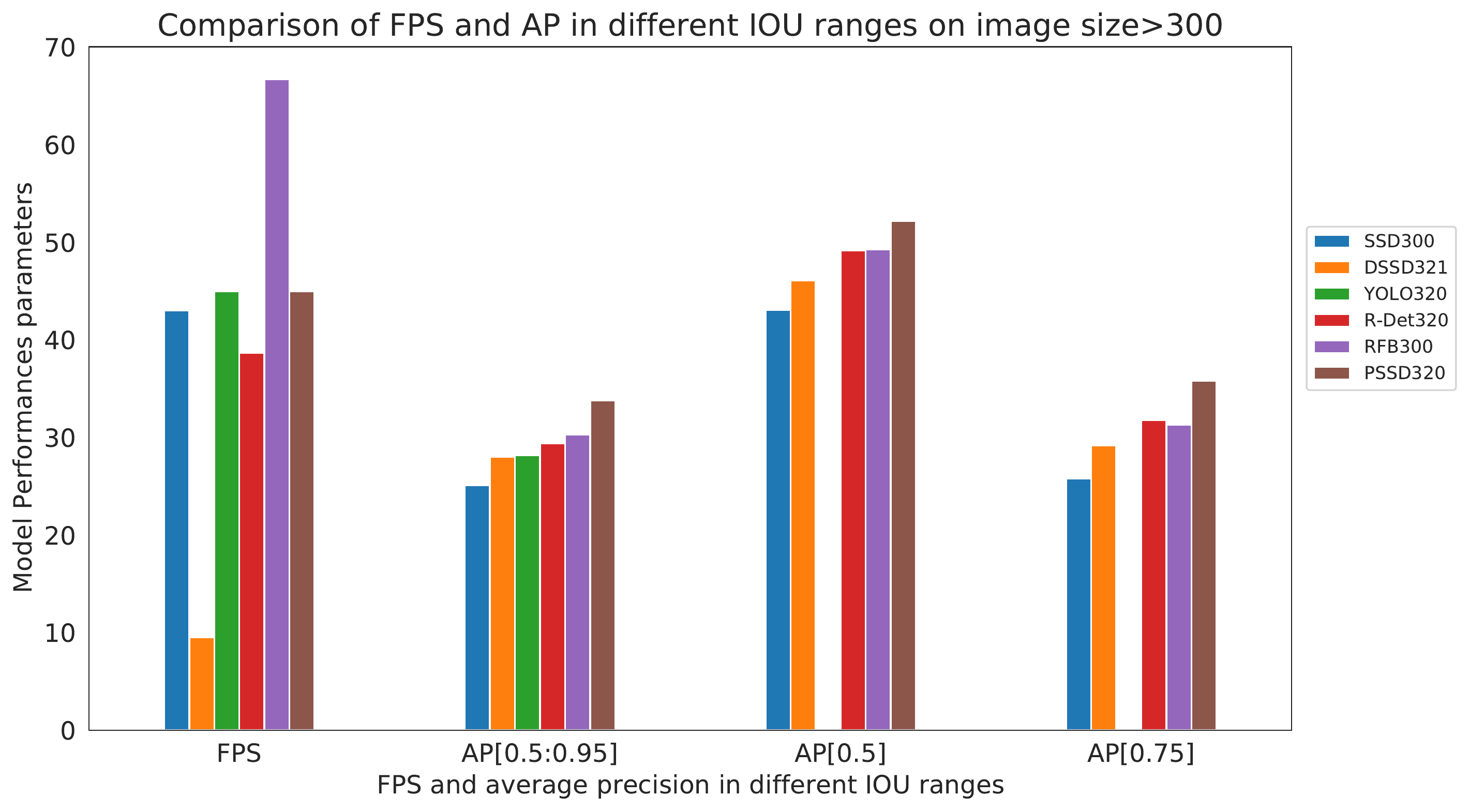}
    \end{minipage}
    }
    \caption{Performance comparison in term of FPS and AP, for different IOU and area using input size more than 300 and less than 321}
    \label{Per_fig_300}
\end{figure}
The detection accuracy comparison between the present PSSD model and other advanced state-of-the-art detectors such as Faster R-CNN \cite{ren2015faster}, Faster R-CNN w FPN \cite{lin2017feature}, R-FCN \cite{dai2016r}, CoupleNet \cite{zhu2017couplenet}, Deformable R-FCN \cite{dai2017deformable}, SSD300 \cite{liu2016ssd}, DSSD321 \cite{fu2017dssd}, YOLO$_{v3}$320 \cite{redmon2018yolov3}, RefineDet320 \cite{zhang2018single}, RFB-Net300 \cite{liu2018receptive}, SSD512 \cite{liu2016ssd}, DSSD513 \cite{fu2017dssd}, YOLO$_{v3}$608 \cite{redmon2018yolov3}, RefineDet512 \cite{zhang2018single}, RFB-Net512 \cite{liu2018receptive} has been considered  for various backbones including VGG16, ResNet-101, Inc-Res-v2, and  DarkNet53 as shown in  Table \ref{table1}.

\begin{figure}
    \centering
    \subcaptionbox{}{
    \begin{minipage}[b]{1.0\linewidth}
    \includegraphics[width=1\linewidth]{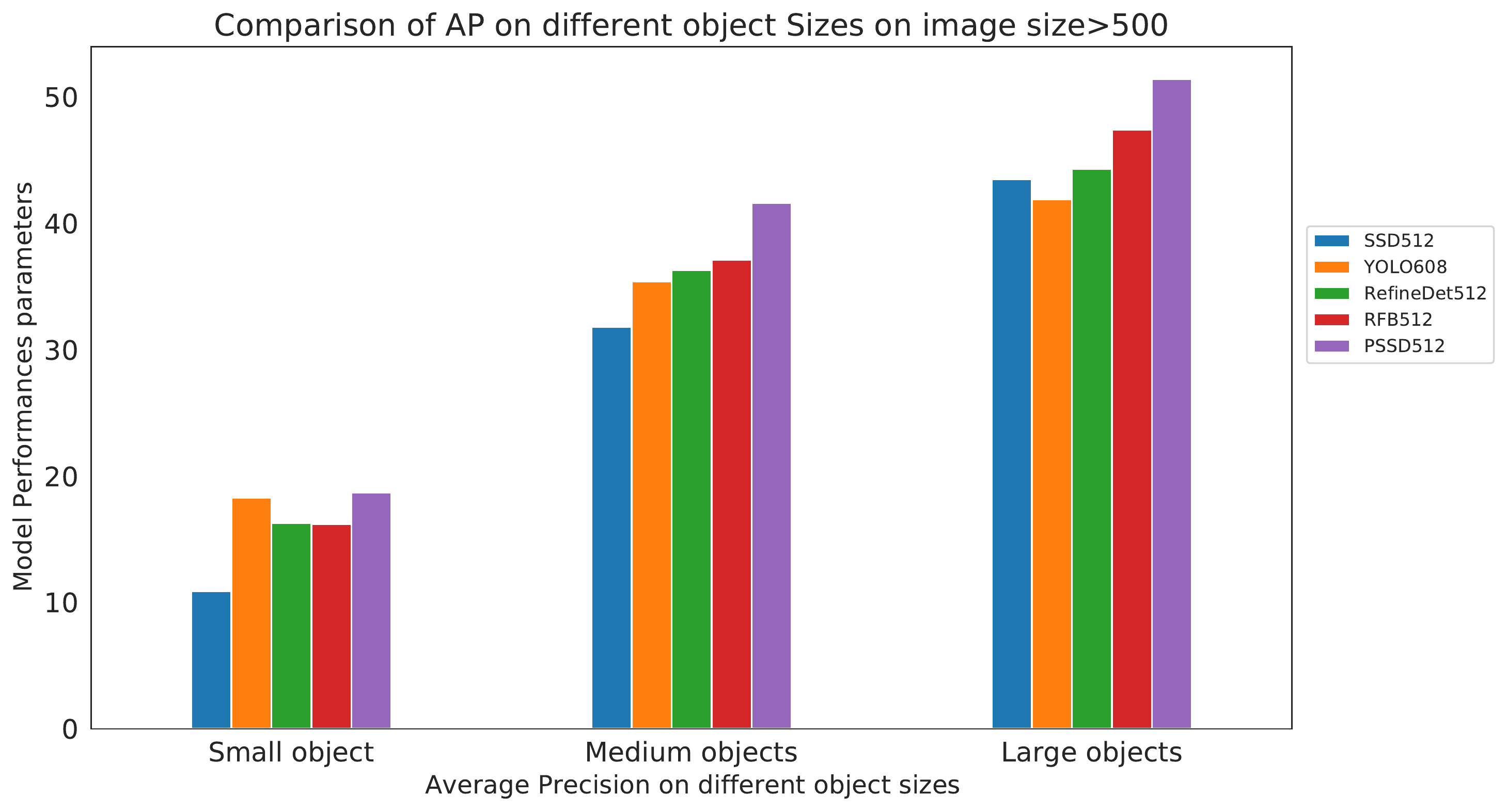}
    \end{minipage}
    }
    \subcaptionbox{}{
    \begin{minipage}[b]{1.0\linewidth}
    \includegraphics[width=1\linewidth]{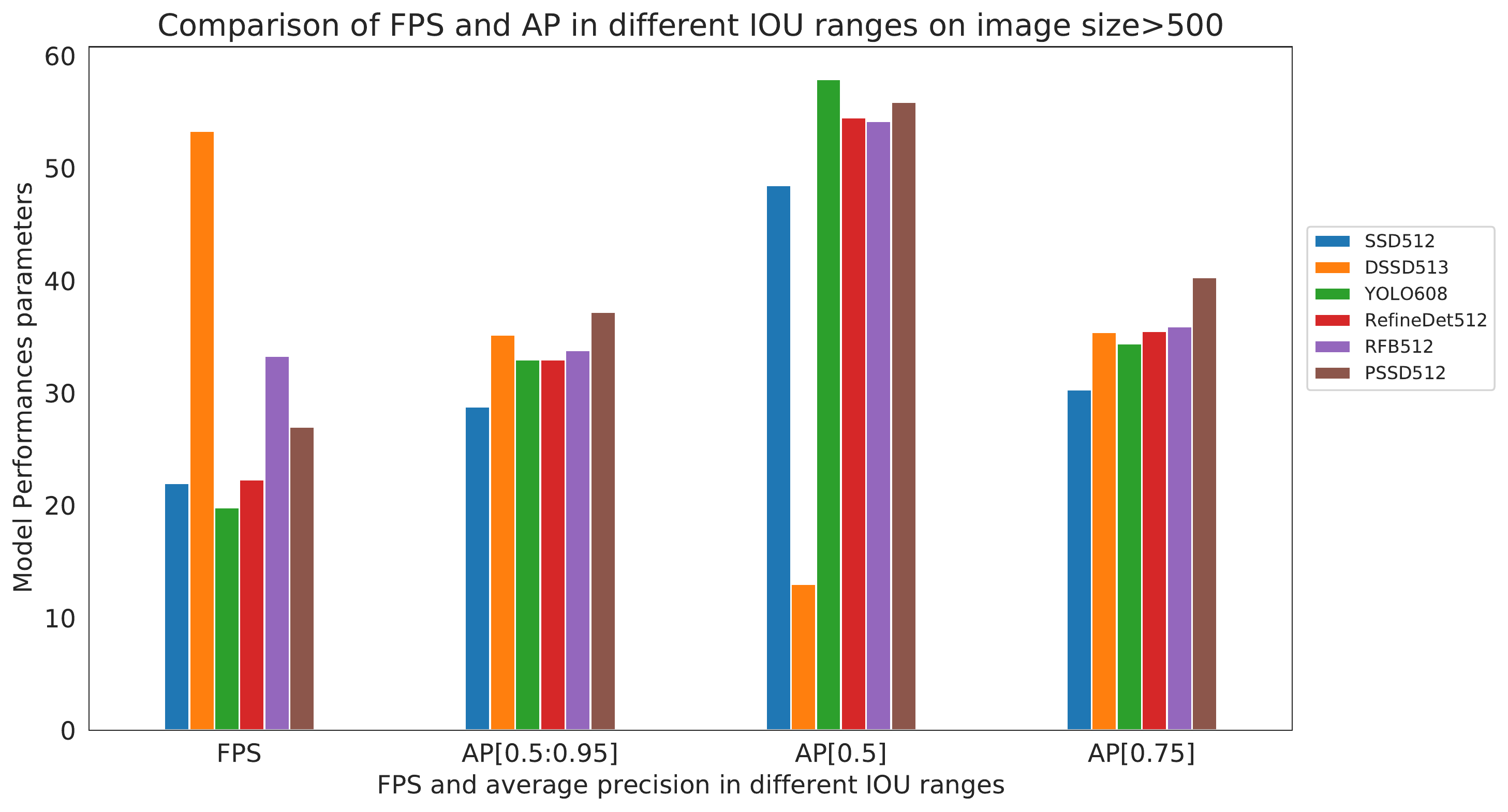}
    \end{minipage}
    }
    \caption{Performance comparison in term of FPS and AP, for different IOU and area using input size more than 500}
    \label{Per_fig_500}
\end{figure}

From the comparison, it is worth noting that the PSSD320 achieves 33.8 mAP, surpassing most object detectors with more powerful backbones and larger input sizes. For example, R-FCN  performs 29.9 mAP at the input size of 1000$\times$600 pix, and DSSD513  performs 33.2 mAP. Even if they use a deeper ResNet-101 \cite{he2016deep}, the mAP is lower than the PSSD320. At the same time, our model retains a powerful inference speed at 45 FPS based on the optimization. Compared with other real-time methods, PSSD has higher precision. For example, RFB-Net300  achieves 30.3 mAP, and RefineDet320  achieves 29.4 mAP. In addition, under the input size of 512 pix, our model also shows a good performance with 37.2 mAP at 27 FPS. Its accuracy and speed  exceed YOLO$_{v3}$608 \cite{redmon2018yolov3} and sophisticated RetinaNet500 \cite{lin2017focal}.
For better clarity in the comparison, we visualize the result in different groups i.e. average precision in different areas such as small, medium, and large as shown in Fig. \ref{Per_fig_300}-(a), it is noticeable that PSSD310 in all area sizes outperformed the performance of the other models. We not only visualize the results from an area perspective but also visualized the results from FPS, and different AP thresholds. As shown in Fig. \ref{Per_fig_300}-(b), PSSD312 has consistently outperformed the previous methods using different AP threshold ranges while PSSD312 has the second-highest FPS among all methods. Like \ref{Per_fig_300},  we do visualization to check the effect of different input sizes, we change input size from 320 to 512. As it is shown in figure \ref{Per_fig_500} (a), PSSD512 with different area objects has outperformed all other methods, while in terms of FPS, performance is dropped but still comparable, and in terms of different AP threshold ranges, it outperformed all other methods except AP [0.5], where Yolo has demonstrated highest performance as shown in Fig. \ref{Per_fig_500}-(b). In summary,  our proposed PSSD has shown massive performance gain in terms of different accuracy parameters and detection speed.    
\\ \\ 
\begin{table}[hb]
\centering
\caption{Average Precision Comparison of  Modified SSD.\newline}
\scalebox{1}{
\begin{tabular}{ccccccc}
\hline  
Method & AP & $AP_{50}$ & $AP_{75}$& $AP_{small}$ & $AP_{medium}$ & $AP_{large}$\\
\hline  
SSD$^{\&}$ & 28.5 & 46.9 & 29.9 & 0.98 & 30.8 & 46.2\\
Modified SSD$^{\&}$ & 28.8 & 46.3 & 30.1 & 10.9 & 31.0 & 45.6\\
SSD$^{\&}$\_FPN & 29.7 & 49.0 & 30.7 & 12.2 & 32.4 & 45.4\\
Modified SSD$^{\&}$\_FPN & 30.5 & 50.0 & 31.9 & 12.7 & 33.8 & 46.1\\
\hline
\end{tabular}
}
 \\ \hspace*{\fill} \\
 
\label{table3}
\end{table}
\\
\\
\\{\bf 5.2 Ablation Study:}
\\
\\
Since our model consists of a modified SSD backbone and multiple sub-modules, we verify the validity of each module extensively. The baseline is a simple detector of the original SSD \cite{liu2016ssd} with a 320$\times$320 input size.
\begin{table}
  \centering
     \caption{Ablation study on PSSD320: The detection results are evaluated on \emph{2017val} set \newline}
  \begin{tabular}{c|ccccccc}
  \hline  
  + Bn-Balancel$_1$ &  & \checkmark & \checkmark & \checkmark & \checkmark \\
  + Two-way FPN &  &  & \checkmark & \checkmark & \checkmark \\
  + RFM &  &  &  & \checkmark & \checkmark \\
  + IOU-guided prediction &  &  &  &  & \checkmark \\
  \hline
  AP & 25.8 & 28.8 & 31.5 & 32.2 & 33.8 \\
  $AP_{50}$ & 44.7 & 46.3 & 51.0 & 51.2 & 52.2 \\
  $AP_{75}$ & 26.8 & 30.1 & 32.9 & 33.7 & 35.8 \\
  $AP_{small}$ & 7.2 & 10.9 & 13.5 & 13.8 & 14.8 \\
  $AP_{medium}$ & 27.4 & 31.0 & 34.7 & 35.7 & 38.5 \\
  $AP_{large}$ & 41.4 & 45.6 & 47.6 & 48.3 & 50.3  \\
  \hline
  \end{tabular}
 \label{table2}
  \end{table}
\\
\\
{\bf 5.2.1  Modified Single-shot Multibox Detector SSD$^\&$ backbone:}
\\
\\
To verify the effectiveness of the  SSD$^\&$ backbone for the feature fusion module, we conduct four experiments with FPN. The results are shown in Table \ref{table3}. First of all, it makes the original SSD mAP increase by 2.7\% that adding BatchNorm and Balance l$_1$ \cite{pang2019libra}. The accuracy of SSD$^\&$ after modifying the extra layers is 28.8\% which is similar to the original SSD (removing the effect of increasing parameters). However, adding the FPN structure to the modified SSD$^\&$ has a better effect that the mAP reaches 30.5\%. The original SSD-FPN is only 29.7\%.In contrast, under the FPN structure, the modified backbone is increased by 0.5\%. This shows new extra layers are more suitable for the feature fusion structure.
\\
\\
{\bf 5.2.2  FEM module:}
\\
\\
The FEM module operation process is divided into two steps. Firstly, the receptive field of basic features is expanded by RFM step by step and then we use Two-way FPN to supplement the more in-depth and shallower information. It can be seen from Table \ref{table2} and Table \ref{table3} that the Two-way FPN is more effective than the original FPN, especially under the large object and the IOU$_{0.75}$ index, the overall reach 31.5\% which is 1\% higher than the original FPN. The addition of RFM further improves the model's mAP to 32.2\% in line with our expectations. In short, benefiting from the FEM module, the feature information of each feature map is more abundant, and the detection effects of both large and small objects are greatly improved.
\\
\\
\begin{figure}
    \centering
    \subcaptionbox{}{
    \begin{minipage}[b]{0.85\linewidth}
    \includegraphics[width=1\linewidth]{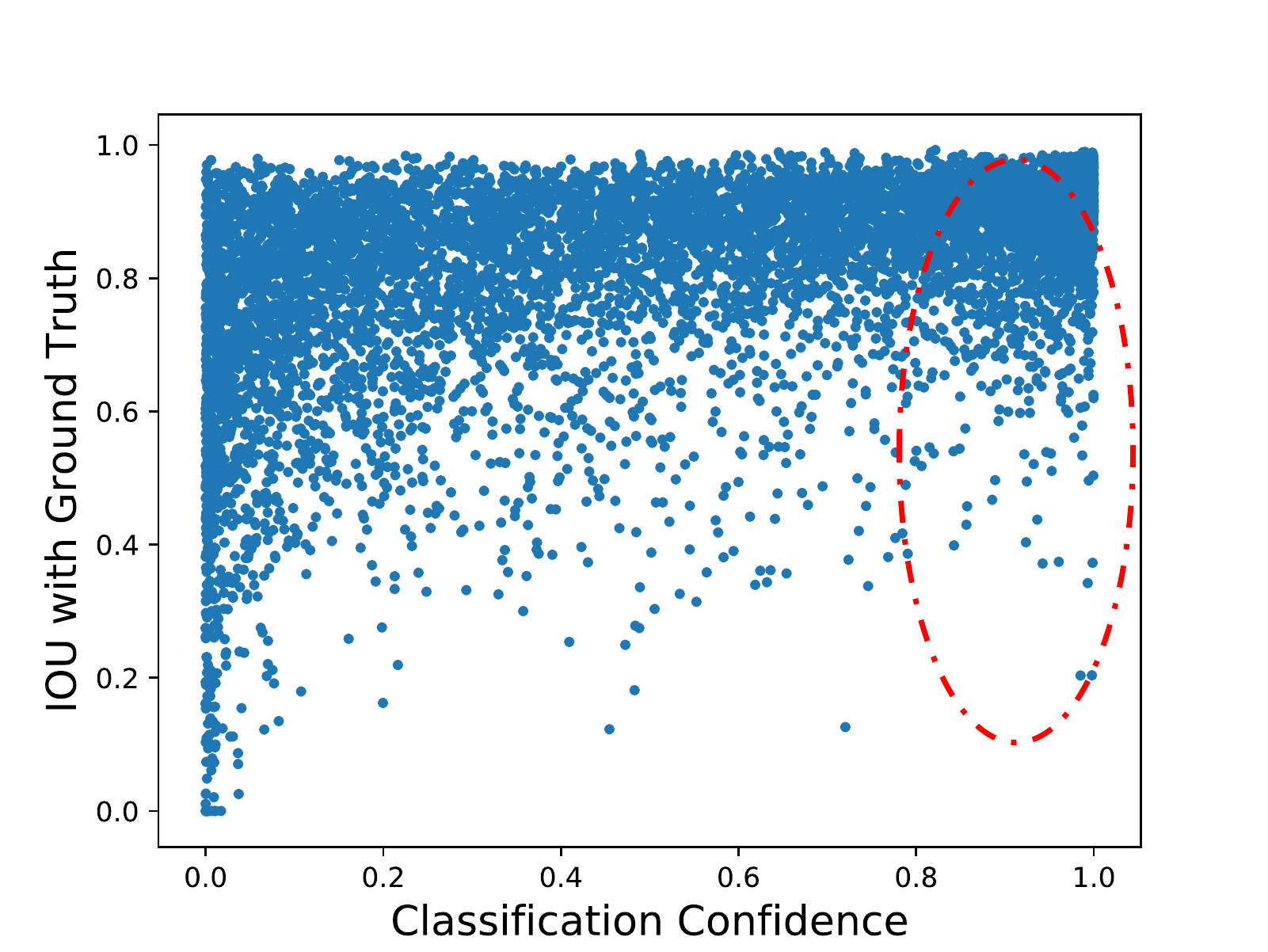}
    \end{minipage}
    }
    \subcaptionbox{}{
    \begin{minipage}[b]{0.85\linewidth}
    \includegraphics[width=1\linewidth]{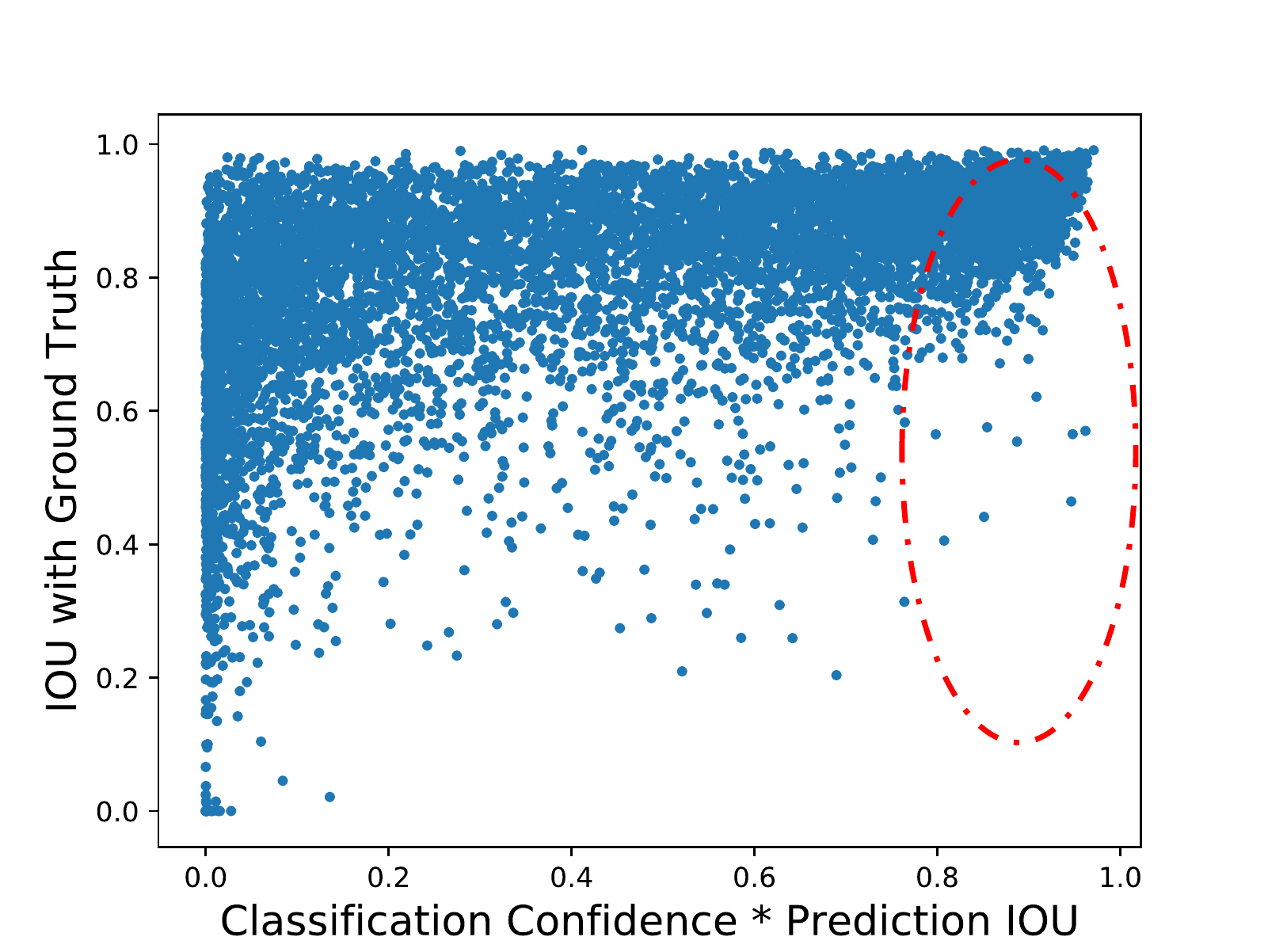}
    \end{minipage}
    }
    \caption{Distribution of NMS scores and real IOUs on MS COCO 2017val set test results. The prediction boxes in the red circle is significantly reduced.}
    \label{fig7}
\end{figure}
{\bf 5.2.3  IOU Prediction structure:}
\\
\\
We predict the IOU between the regression boxes and their ground truth with very little overhead and use them in classification loss and inference. Overall, the IOU-guided prediction structure obtains effective results and improves mAP by 1.6\% as shown in Table \ref{table2}. In addition, we verify the validity of R$_{IOU}$ loss and CEJI loss. As shown in  Table \ref{table4}, Comparing R$_{IOU}$ loss with l$_2$ loss, the new loss has a better effect with an increase of 0.5\%. At the same time, CEJI loss has increased by 0.8\% compared to CE loss. It means that the two new loss functions can both better optimize the model. As can be seen from Fig. \ref{fig7}, the number of prediction boxes that have a high score but low IOU$_{tar}$ is significantly less, so it is more reasonable to use the new score to conduct overlapping boxes filtration.
\begin{table}
\centering
\caption{Comparisons of different loss function settings.\newline}
\scalebox{1}{
\begin{tabular}{ccccccc}
\hline  
 loss & AP & $AP_{50}$ & $AP_{75}$ & $AP_{small}$ & $AP_{medium}$ & $AP_{large}$\\
\hline  
 L$_2$+CEJI & 33.3 & 51.9 & 35.5 & 14.9 & 37.8 & 50.2\\
 R$_{iou}$+CE & 33.0 & 51.7 & 34.7 & 15.1 & 36.5 & 49.5\\
 R$_{iou}$+CEJI & 33.8 & 52.2 & 35.8 & 14.8 & 38.5 & 50.3\\
\hline
\end{tabular}
 }
\label{table4}
\end{table}
\\
\\
{\bf 5.3 Performance evaluation on Pascal VOC 2007}
\\
\\
To further validate PSSD, we conduct experiments on a relatively small Pascal VOC dataset. Specifically, we train our model on the union of 2007 \emph{trainval} set and 2012 \emph{trainval} set. This training set contains 20 categories and 16555 images and then we test the result on VOC \emph{2007test} set which has 5k images. We set the batch size to 32, using the default boxes of 0.06, 0.15, 0.33, 0.51, 0.69, 0.87, 1.05 ratio, and the weight decay of 0.0005, the momentum of 0.9. The initial learning rate is set to 0.002, attenuated by 10 times at 160th and 200th epochs. Finally, the training stops at the 240th epoch.
Table \ref{table5} shows the comparison of our experimental results with other state-of-the-art (SOTA) models. The performance of PSSD320 is better than that of RefineDet320 \cite{zhang2018single}, YOLO$_{v3}$320 \cite{redmon2018yolov3} and RFB-Net300 \cite{liu2018receptive}. Its mAP is 81.28\% which exceeds the original SSD300 \cite{liu2016ssd}  by 4\% while maintaining the 66FPS real-time speed. The PSSD512 achieves 82.82\% mAP at the speed of 40 FPS, which is superior to most Single-stage and Two-stage object detection systems. It can be seen that our model also shows a good performance on small-scale datasets.
\begin{table}
\caption{Detection accuracy comparisons on Pascal VOC 2007test set. \newline}
\label{table5}
\centering
\scalebox{1}{
\begin{tabular}{cccccccccccccccccccccccc}
\hline  
Method & Backbone & FPS & mAP\\
\hline  
Faster R-CNN  & VGG16 & 7  & 73.2\\
Faster R-CNN w FPN   & ResNet-101 & 5  & 76.4\\
R-FCN  & ResNet-101 & 9  & 80.5\\
Deformable R-FCN  & ResNet-101 &9  & 82.6\\
SSD300  & VGG16 & 120  & 77.2\\
DSSD321  & ResNet-101 & 9.5  & 78.6\\
RefineDet320  & VGG16 & 83  & 80.5\\
RFB-Net300  & VGG16 & 40.3  & 80.0\\
\textbf{PSSD320} & VGG16 & \textbf{66} & \textbf{81.28}\\
SSD512  & VGG16 & 50  & 79.8\\
DSSD513  & ResNet-101 & 5.5  & 81.5\\
RefineDet512  & VGG16 & 83  & 80.5\\
RFB-Net512 & VGG16 & 24.1  & 81.8\\
\textbf{PSSD512} & VGG16 & \textbf{40} & \textbf{82.82}\\
\hline
\end{tabular}
 }
 \\ \hspace*{\fill} \\
 
\end{table}
\\
\\
{\bf 6. Discussion}\label{sec:discussion}
\\
\\
{\bf 6.1  Trade-off between speed and accuracy : }
\\
\\
 Our model has a better trade-off between speed and accuracy. PSSD has better accuracy and speed than RetinaNet500 \cite{lin2017focal} and yolo$_{v3}$608 \cite{redmon2018yolov3}. Meanwhile, PSSD has higher accuracy than RFB-Net \cite{liu2018receptive} and RefineDet \cite{zhang2018single}, it is due to our efficient method optimization. On the one hand, the FEM module makes each feature map used by predictors get richer semantics and local information to achieve better prediction results. On the other hand, in the NMS algorithm, according to to multiply the classification confidence by the predicted IOU as the final scores, the boxes with high classification confidence but inaccurate regression are more attenuated to keep more accurate prediction boxes as much as possible. The two aspects complement each other to make our model obtain more accurate prediction boxes. At the same time, the proposal of R$_{iou}$ loss and CEJI loss enable the model to learn better. Considering the inference speed, the design of each module is as lightweight as possible.
\\
\\
\begin{figure}[ht]
\begin{subfigure}{.5\textwidth}
  \centering
  \includegraphics[width=.8\linewidth]{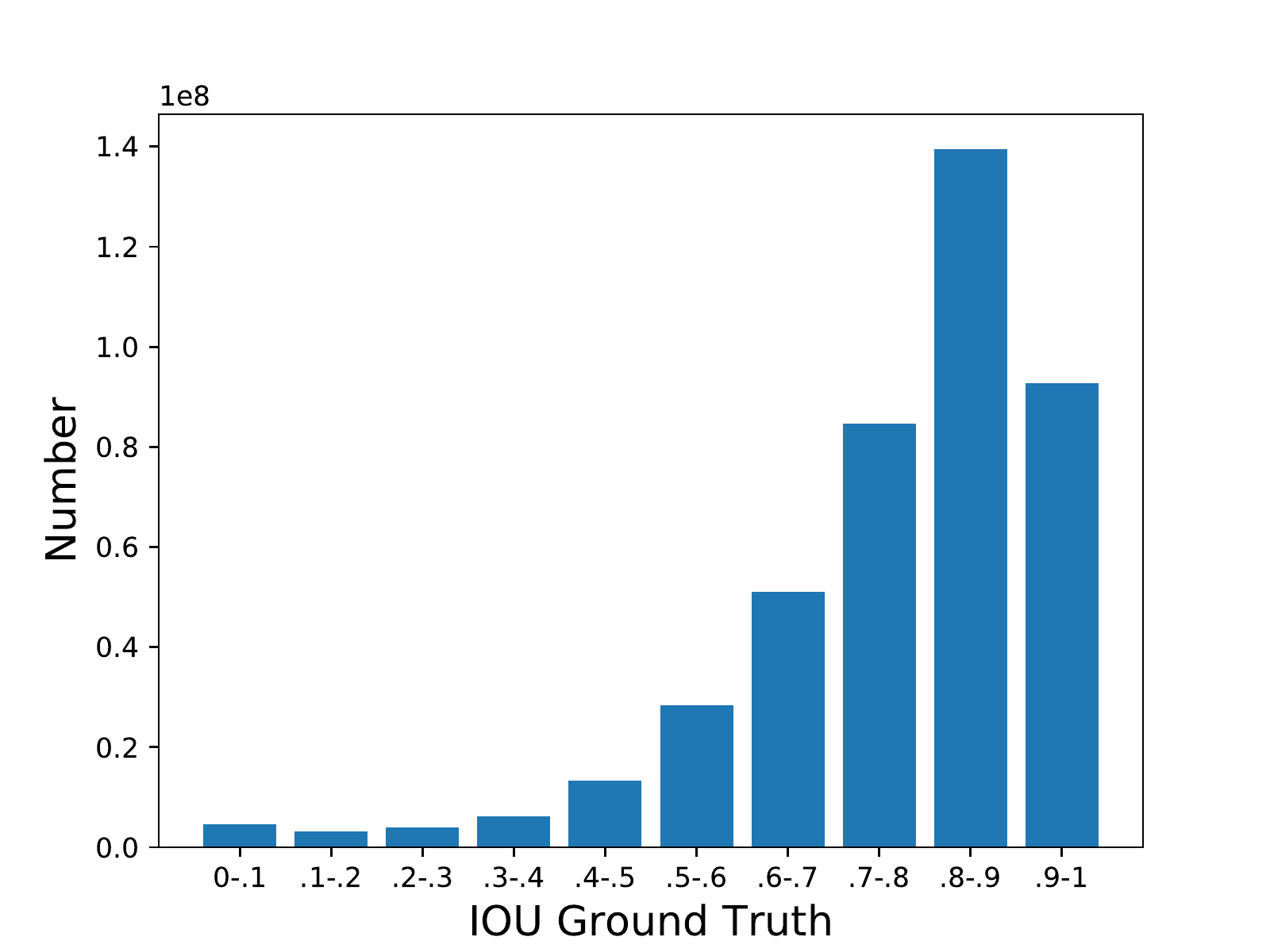}  
  \caption{0-150 epochs}
\end{subfigure}
\begin{subfigure}{.5\textwidth}
  \centering
  \includegraphics[width=.8\linewidth]{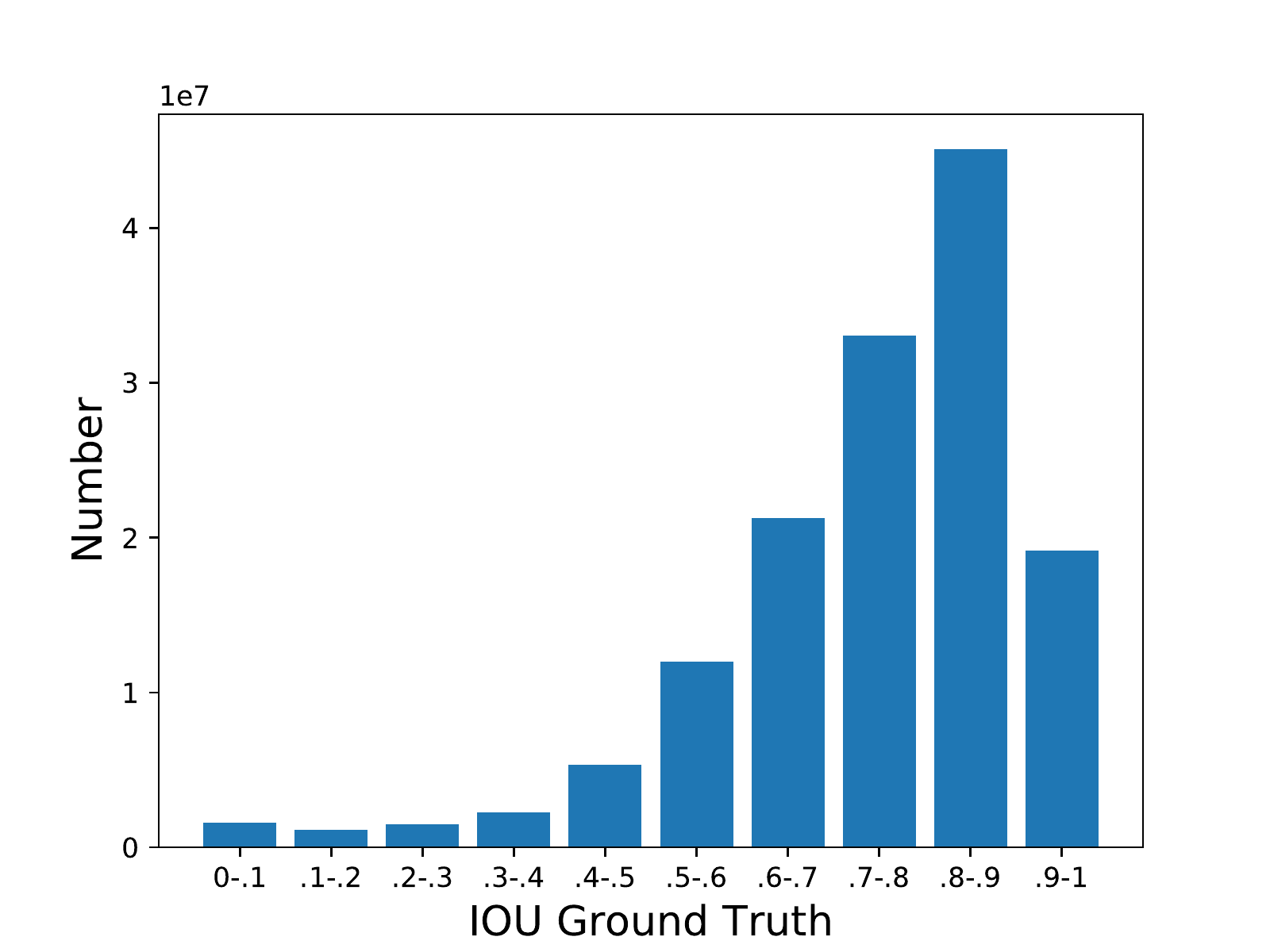}  
  \caption{0-50 epochs}
\end{subfigure}
\newline
\begin{subfigure}{.5\textwidth}
  \centering
  \includegraphics[width=.8\linewidth]{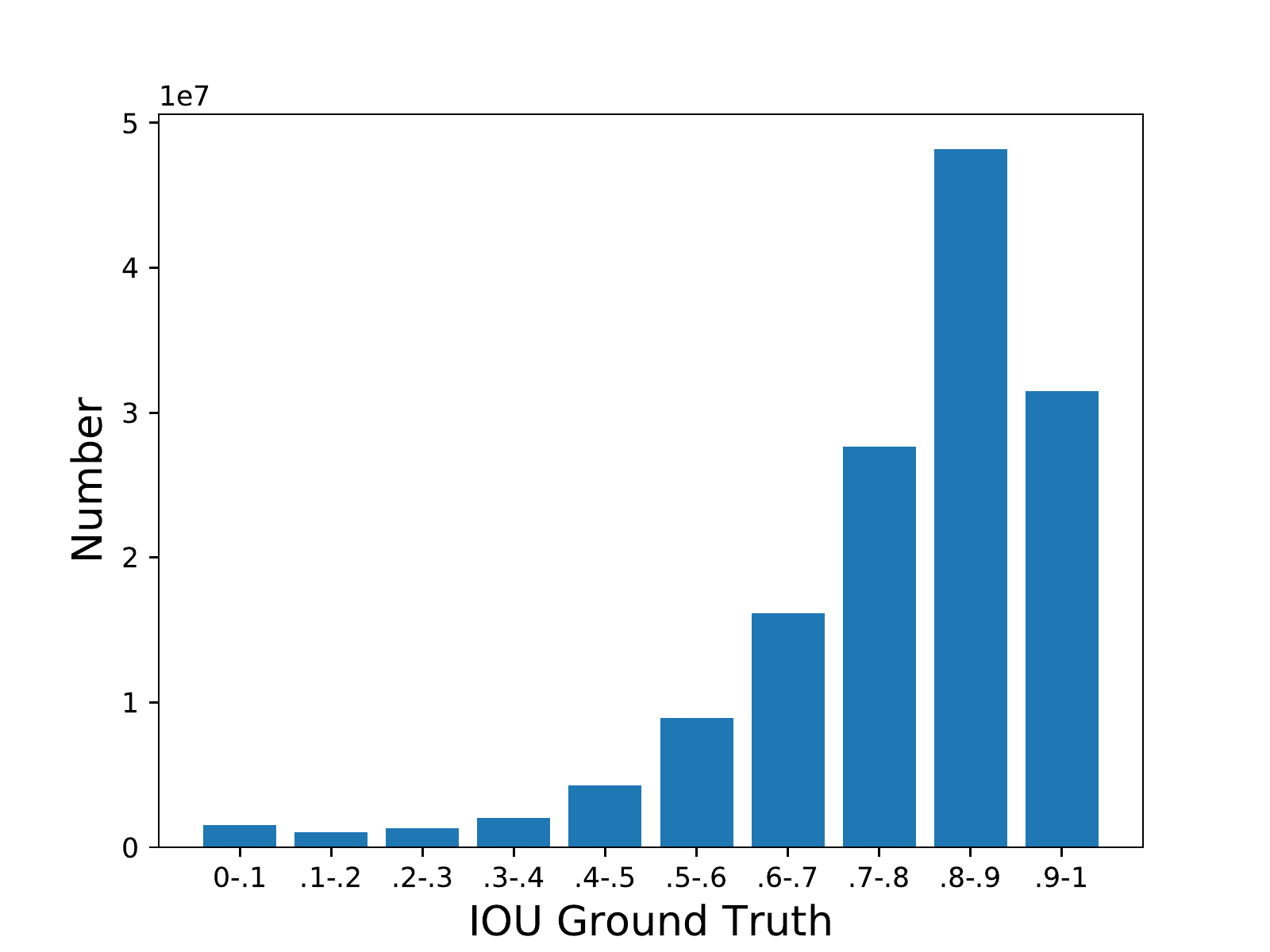}  
  \caption{50-100 epochs}
\end{subfigure}
\begin{subfigure}{.5\textwidth}
  \centering
  \includegraphics[width=.8\linewidth]{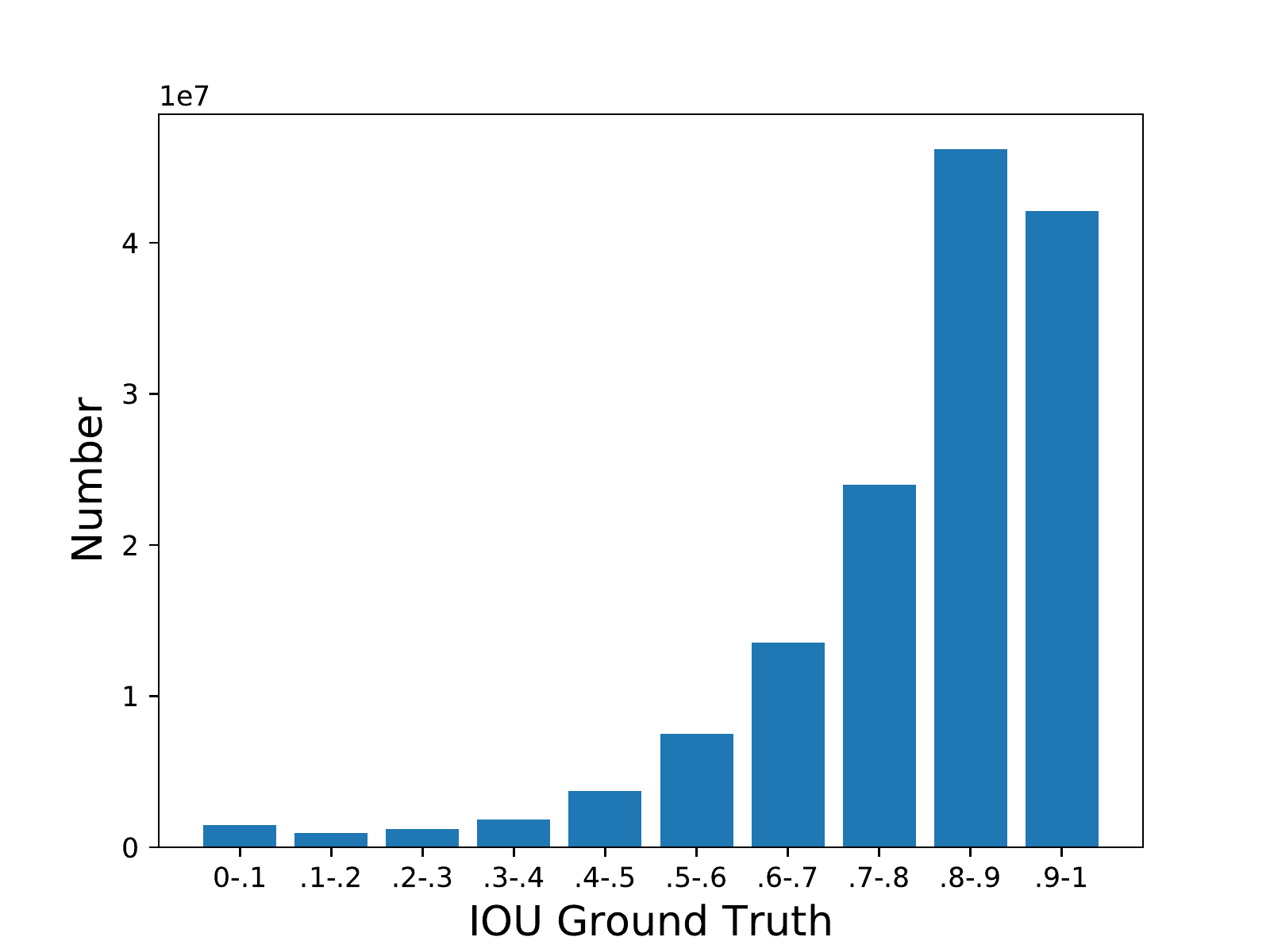}  
  \caption{100-150 epochs}
\end{subfigure}
\caption{Distribution of $IOU_{tar}$ at different training stages: (1) Overall unevenness; (2) the IOU samples above 0.9 is significantly increasing as the training deepens.}
\label{fig:fig8}
\end{figure}
\\
\\
{\bf 6.2 Existing problems : }
\\
\\
A significant problem with Single-stage architecture is the class imbalance between positive and negative samples. Similarly, as shown in Figure \ref{fig:fig8}, this problem is more severe in IOU prediction. There are two situations during the training. Firstly, the distribution of IOU samples at each interval is uneven. Secondly, the distribution of IOU samples in the training process is constantly changing. These two points lead to a severely imbalanced IOU sample. Therefore, the effect of IOU prediction is unstable, the overall value is high and the accuracy is low. However, in this case, it can still improve by 1.6\%, which indicates the prospect of the method, and we have a plan to optimize this problem more in future study.
\\
\\
{\bf 7. Conclusion}
\\
\\
In this paper, we address two issues for inaccurate prediction boxes of the Single-stage architecture and proposed new architecture named PSSD and optimize it based on the Single-stage SSD model. Firstly, we introduce the extra layers in the SSD backbone to make it more suitable for high representative feature fusion. Secondly, we use the FEM module to make information on each prediction feature richer and more beneficial to the classification and regression task of each scale. Finally, we design the IOU-guided prediction structure to make the model optimize better. Our proposed architecture showed impressive performance on two popular and large datasets i.e. MS COCO and Pascal VOC and outperformed the previous SOTA methods. Its accuracy even surpasses some of the complex top detectors while maintaining an exciting speed of inference.
\newline  
{\bf Acknowledgement} This work was partially supported by the Key Program of the National Natural Science Foundation of China under Grant No. 61932014 and the Research and Development Projects of Applied Technology of Inner Mongolia Autonomous Region, China under Grant No. 201802005. \newline
 {\bf Conflict of interest} The authors declare that they have no conflict of interest.  \newline 
  {\bf Data availability} The data that support the findings of this study are available to public. Cocodataset link  : https://cocodataset.org/; PASCAL VOC dataset link: \\  http://host.robots.ox.ac.uk/pascal/VOC/

	%
	%
	%
	%
	%
	\newpage
	\bibliographystyle{elsarticle-num}
	
	\bibliography{2.SSD_ref} 
	

\end{document}